\documentclass{article} %
\usepackage{colm2024_conference}

\usepackage{microtype}
\usepackage{hyperref}
\usepackage{url}
\usepackage{booktabs}
\definecolor{darkblue}{rgb}{0, 0, 0.5}
\hypersetup{colorlinks=true, citecolor=darkblue, linkcolor=darkblue, urlcolor=darkblue}

\usepackage[T1]{fontenc}
\usepackage{graphicx}
\usepackage{enumerate}
\usepackage{amsmath}
\usepackage{amsthm}
\usepackage{array}
\usepackage{tabularx}
\usepackage{cleveref}
\usepackage{multirow}
\usepackage{caption}
\usepackage{subcaption}

\renewcommand{\paragraph}[1]{\textbf{#1}\hspace{.2cm}}
\newtheorem{theorem}{Theorem}[section]
\newtheorem{definition}[theorem]{Definition}
\newtheorem{proposition}[theorem]{Proposition}

\title{Latent Causal Probing: \\
A Formal Perspective on Probing with Causal Models of Data}

\author{{Charles Jin \& Martin Rinard} \\
MIT CSAIL \\
Cambridge, MA 02142, USA \\
\texttt{\{ccj,rinard\}@csail.mit.edu} \\
}

\colmfinalcopy %
\begin{document}

\maketitle

\begin{abstract}
As language models (LMs) deliver increasing performance on a range of NLP tasks, \emph{probing classifiers} have become an indispensable technique in the effort to better understand their inner workings.
A typical setup involves (1) defining an auxiliary task consisting of a dataset of text annotated with labels, then (2) supervising small classifiers to predict the labels from the representations of a pretrained LM as it processed the dataset. A high probing accuracy is interpreted as evidence that the LM has learned to perform the auxiliary task as an unsupervised byproduct of its original pretraining objective.
Despite the widespread usage of probes, however, the robust design and analysis of probing experiments remains a challenge. We develop a formal perspective on probing using \emph{structural causal models} (SCM). Specifically, given an SCM which explains the distribution of tokens observed during training, we frame the central hypothesis as whether the LM has learned to represent the latent variables of the SCM. Empirically, we extend a recent study of LMs in the context of a synthetic grid-world navigation task, where having an exact model of the underlying causal structure allows us to draw strong inferences from the result of probing experiments. Our techniques provide robust empirical evidence for the ability of LMs to induce the latent concepts underlying text.
\end{abstract}

\section{Introduction}

As large LMs pretrained on massive amounts of unlabeled text
continue to reach new heights in NLP tasks (and beyond), the question of what kinds of information such models encode about their training data remains a topic of intense discussion and research. One prominent technique is to supervise small \emph{probing classifiers} to extract some linguistically relevant property from the representations of the pretrained LM \citep{shi2016does,adi2017finegrained,alain2018understanding}, with the intuition being that the success of the probe reveals the LM has, in fact, learned to encode the property of interest as a byproduct of its training.

Despite their widespread usage, however, probes themselves are also an active area of research, with a number of interconnected open questions in the design and interpretation of probing experiments \citep{belinkov-2022-probing}, including:

\paragraph{(Q1) Control and interpretation.}
Given that the probe itself is directly supervised to perform the auxiliary task, the observed outcomes could depend not only on the information inherently encoded in the LM but also the ability of the probe to extract the information itself. For instance, researchers have found that training probes to predict \emph{randomized} labels can often yield comparably high accuracies on certain tasks, calling into question the significance of prior results \citep{hewitt2019designing}. As a result, drawing robust conclusions from the classification accuracy of a probe remains up for debate.

\paragraph{(Q2) Classifier selection and training.}
To combat the risk of measuring the probe's capacity to learn the auxiliary task, researchers often limit probes to low capacity architectures such as linear classifiers \citep{hall-maudslay-etal-2020-tale}. However, other works have countered with evidence that LMs encode more complex concepts using non-linear representations, which can only be accurately measured using higher capacity classifiers \citep{10.1162/tacl_a_00254,li2022emergent}.
A related question which has received little attention is how the training procedure itself (e.g., optimizer selection, training hyperparameters, auxiliary dataset size) interacts with the outcome of the probing experiment.

\paragraph{(Q3) Auxiliary task design.}
Finally, as large, pretrained LMs have progressed from producing human-like text to exhibiting increasingly ``intelligent'' behaviors such as reasoning and in-context learning \citep{brown2020language}, there is an emerging need to better understand the limitations and capabilities of LMs along dimensions such world knowledge and theory of mind. These domains present a distinct set of challenges compared to traditional linguistic tasks such as part-of-speech tagging and dependency parsing.

The theoretical section of this paper develops a formal perspective on probing using the language of \emph{structural causal models} (SCM). Specifically, given a causal model which explains the distribution of tokens observed during training, we pose the central hypothesis as determining whether the LM has learned to represent the \emph{latent variables} of the SCM: concepts that explain how the text was generated, but are never directly observed during training. We then introduce probes as a means of empirically testing such hypotheses, by extracting the value of the latent concepts given only the LM representations as input. Our setting naturally captures broader questions about the inductive bias of LMs trained solely on text, and the latent concepts they acquire over the course of training (\textbf{Q3}).

Next, by extending the SCM beyond the data generation process to cover the training of the LM (unsupervised) and probe (supervised), we further show that \textbf{Q1} and \textbf{Q2} can be understood as the mediating and moderating effects of the probe, respectively. We propose a general technique based on \emph{causal mediation analysis} which isolates the causal path through the LM while excluding the probe's influence. Our analysis yields clear, testable conditions for accepting or rejecting our hypotheses based on a probing experiment's outcomes.

Finally, we conduct an empirical study that extends the experimental setting introduced by \citet{jin2023evidence}, who use probes to quantify the extent to which LMs are capable of learning ``meaning'' from text, as operationalized by the semantics of a synthetic programming language for grid-world navigation. By leveraging the proposed \textbf{latent causal probing} framework, our experiments allow us to draw precise conclusions about the \emph{causal relationship} between the latent dynamics that generated the training data and what is learned by the LM. In particular, we find evidence that (1) the LM has, in fact, learned to represent the latent variables corresponding to the underlying semantics of the language, and (2) the LM representations exhibit an inductive bias that generalizes to novel action sequences. Our study marks the first rigorous empirical evaluation of the hypothesis that \emph{language models are latent concept learners}, revealing intriguing insights into how language models might acquire an understanding of language.

\section{Structural causal models of text}

This section introduces the setting of our framework for probing, which is based on the idea that the text used to train LMs may exhibit latent causal structure; we formalize these concepts using the approach of structural causal models.

\subsection{Background: structural causal models}

\begin{figure}[tb]
 \centering
 \begin{subfigure}[b]{0.45\textwidth}
     \centering
     \includegraphics[scale=1.0]{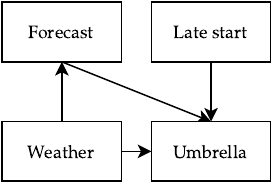}
     \caption{The original SCM.}
     \label{fig:umbrella}
 \end{subfigure}
 \hfill
 \begin{subfigure}[b]{0.45\textwidth}
     \centering
     \includegraphics[scale=1.0]{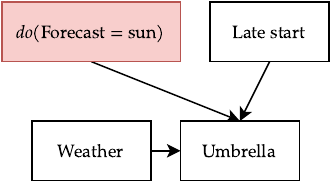}
     \caption{Intervening on the forecast.}
     \label{fig:umbrella_intervene}
 \end{subfigure}
\caption{An SCM for bringing an umbrella to work.}
\end{figure}

Structural causal models are graphical models which represent causal relationships in a data generation process as a directed graphical model \citep{pearl2000models}. %
We illustrate the key concepts with an example; we refer the reader to \citet{pearl2010introduction} for a more comprehensive overview. Suppose that we are interested in the effect the weather has on employees bringing an umbrella to work. In this case, we may hypothesize a SCM like the one in \Cref{fig:umbrella}.
Each node represents a different random variable: the weather, the weather forecast, whether the employee's morning gets off to a late start, and whether the employee brings an umbrella to work. Nodes without parents are \emph{exogenous} variables, whose causes are left unexplained; they are often used to model nature, randomness, or other aspects of physical reality, such as genetic or environmental factors. The exogenous variables in \Cref{fig:umbrella} are the weather and having a late start.
Nodes with a parent indicate the possibility of a causal relationship, e.g., the edge from weather to forecast indicates that the weather might influence the forecast. In particular, every \emph{missing} edge in the SCM asserts the lack of a causal relationship. A standard assumption of causal analysis is that the underlying causal graph is Markovian (or acyclic); we adopt this assumption as well.

\paragraph{Mediators and moderators.} Returning now to our original question of how the weather affects employees bringing an umbrella to work, we note the SCM hypothesizes 3 possible causes: the weather, the weather forecast, and having a late start. The forecast is a \emph{mediator} because total causal effect of the weather on umbrella in partially transferred by the \emph{path-specific effect} over the weather-forecast-umbrella pathway \citep{avin2005identifiability,imai2010general}.
A natural question is how much the forecast is responsible for the increase in likelihood that an employee brings an umbrella to work when, for instance, the weather changes from sunny to rainy. One answer is given by the \emph{necessary indirect effect}, which quantifies how much the presence of the causal path through the mediator contributes to the total measured effect \citep{Weinberger2019}:
\begin{multline*}
    \text{NIE}_{\text{rain}, \text{sun}}(\text{Forecast}) =
    \mathbb{E}\big[\text{Umbrella}
    \mid \text{Weather} = \text{rain}\big] \\ - \mathbb{E}\big[\text{Umbrella} \mid \text{Weather} = \text{rain}, do(\text{Forecast} = \text{sun})\big],
\end{multline*}
where $do(\text{Forecast} = \text{sun})$ is a \emph{causal intervention}. The intervention can be conceptualized as forcing the weather station to forecast sun regardless of the weather, thereby severing the weather-forecast-umbrella pathway. \Cref{fig:umbrella_intervene} depicts the SCM post-intervention.

The late start variable is not a mediator of the weather-umbrella causal effect (because there is no path from the weather to umbrella that passes through it), but it could still be a \emph{moderator}: variables that do not directly mediate a causal effect, but affect the strength (and possibly direction) of another causal path \citep{baron1986moderator}. For instance, the forecast's effect on whether an employee brings an umbrella (i.e., the NIE) might be lower if the employee has a late start and rushes out the door without checking the forecast.

\subsection{Case study: causal structure in programming languages}

\begin{figure}
\begin{center}
\includegraphics[scale=.85]{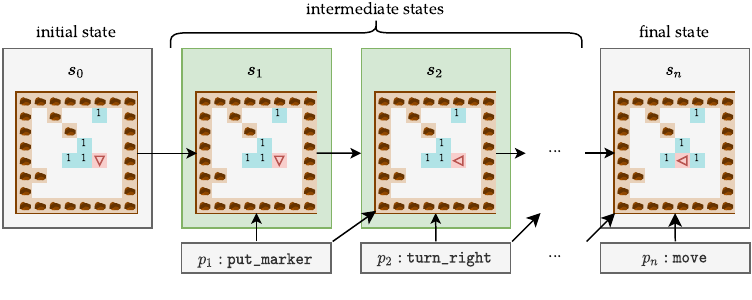}
\end{center}
\caption{An SCM of the data generation process for the grid world corpus. The exogenous variables are the initial state and the actions; latent variables are green and observed variables are gray. The training corpus consists of programs of length between 6 and 10.}
\label{fig:exp_SCM}
\end{figure}

\citet{jin2023evidence}
propose an experiment to study whether LMs are able to ground a sequence of actions into a sequence of states, having only seen instances of the initial and final state during training. Specifically, they train a 350M parameter Transformer \citep{vaswani2017attention} on a corpus of \emph{specification-program} examples using a standard causal language modeling objective. The programs are strings in a grid-world navigation language with 5 actions (\texttt{move}, \texttt{turn\_right}, \texttt{turn\_left}, \texttt{put\_marker}, \texttt{pick\_marker}), sampled uniformly between lengths 6 and 10, inclusive. The specifications consist of the initial and final state, which are 8x8 grids. Executing the program navigates a single robot in the initial state to the final state. We refer to \citet{jin2023evidence} for more details about the language.

\Cref{fig:exp_SCM} displays an SCM of the data generation process (along with an example assignment of values to each variable). The exogeneous variables are the initial state and the program actions. Each action produces a latent state (green), save for the last action, which is observed as the final state. A training sample consists of the sequence: $s_0, s_n, p_1, \ldots, p_n$, where each grid world is converted to text by scanning in row order, with one token per entry.

Consider now modeling a distribution of text drawn from this SCM. In particular, for each sample $x$ there is an assignment $e$ to the exogenous variables in the SCM $M$ such that $M(e) = x$. One strategy would be to learn a model of the SCM depicted in \Cref{fig:exp_SCM}, and integrate the latent variables during inference. For instance, knowing that the robot is one space away from $s_n$ in $s_{n-1}$ could help a learner predict $p_n = \texttt{move}$.

More generally, given observations generated according to some unknown causal mechanism,
a learner could propose various SCMs of the underlying causal mechanism consistent with the observations, then use these SCMs to inform future predictions, an approach known as causal learning \citep{scholkopf2021toward}.
A major challenge in the foregoing approach is the problem of \emph{latent variable induction}, or inferring the latent variables over which candidate SCMs are to be defined.
In this work, we focus on the causal structure of programming languages, where the underlying causal dynamics are governed by a precise \emph{formal semantics}, and the latent variables are given by program states. Having formally defined semantics and latent variables enables us interpret the results of our probing experiments in an unambiguous way; we refer the reader to \citet{sloman2005,feder2022causal} for surveys of causal structure in natural language.

\section{Latent causal probing}
\label{sec:probing}

We present \textbf{latent causal probing}, a formal framework for empirically testing the hypothesis
\begin{center}
\emph{Language models are latent concept learners}.
\end{center}
At a high level, given an SCM that models the training data as the observed variables, we probe the LM for representations of the latent variables of the SCM. Our main insight, as illustrated in \Cref{fig:exp_SCM}, is that knowing the latent value of $s_{n-1}$ could help predict the observed value of $p_n$; hence, an LM trained to predict $p_n$ might eventually induce the existence of the latent variable $s_{n-1}$. %

\subsection{Probing for latent concepts}

We begin by defining the auxiliary task and dataset for probing. Fix some structural causal model $M$, and let $v_M$ be the latent variable of interest. Given some text $x$, we use $v_M(x)$ to denote the value of the latent variable in the SCM of text $x$. For instance, the value of $v_M = s_1$ in the sample $x$ from \Cref{fig:exp_SCM} is the grid depicted in the $s_1$ node. We assume that the value of each latent variable is uniquely determined by $x$ and $M$.

Given a language model $LM$ with parameters $\theta$, we denote an arbitrary representation function as $LM(x; \theta)$. The auxiliary dataset consists of input features $\{LM(x; \theta) \mid x \in X\}$ and labels $\{v_M(x) \mid x \in D\}$, where $D = \{x_i\}_{i=1}^N$ is a corpus of text. We then split $D$ into two auxiliary datasets: one for \textbf{calibration} and one for \textbf{measurement}. The probe is trained to predict $v_M(x)$ given $LM(x; \theta)$ on the calibration data, and the accuracy is taken over the measurement data. We next discuss the design and interpretation of these two datasets.

\paragraph{Bound vs. free latent variable outcomes.}
In general, there may exist several causal dynamics that explain the data equally well. For instance, the following dynamics could also generate the data in \Cref{fig:exp_SCM}:
\begin{enumerate}[\hspace{55pt}(P1)]
\item[\texttt{put\_marker}] Jump to a random location.
\item[\texttt{turn\_right}] Return to the last position, put a marker, then turn right.
\end{enumerate}
These dynamics assign a different value to $s_1$, but explain the observed variables equally well. Assuming the training corpus consists entirely of this single example, it would be impossible to distinguish between $M$ and $M'$ on the basis of data alone. In other words:
\begin{enumerate}
\item $M$ and $M'$ share the same set of set of latent, observed, and exogenous variables;
\item $M$ and $M'$ agree on the observed data; and
\item there exists an assignment $e$ to the exogenous variables such that $v_M(x) \ne v_{M'}(x')$ for $x = M(e)$ and $x' = M'(e)$.
\end{enumerate}
In this case, we say that the latent variable $v$ is \textbf{free} over the assignment $e$. More generally, given a hypothesis class $\mathcal{M}$ of SCMs over the same set of variables, denote the LM training data as $D_\text{train}$ and define $\mathcal{M} |_\text{train}$ to be the subset of SCMs that generate $D_\text{train}$. The free latent variable outcomes consist of pairs of latent variables and assignments $(v, e)$ such that there exist $M, M' \in \mathcal{M} |_\text{train}$ where $v_M(M(e)) \ne v_{M'}(M'(e))$. Any latent variable outcome $(v, e)$ which is not free is \textbf{bound}, i.e., the training data $D_\text{train}$ fully specifies the outcome of $v$ on the assignment $e$, given the hypothesis class $\mathcal{M}$.

\begin{table}[tb]
\centering
\begin{tabular}{ccc}
\toprule
calibration & \multicolumn{2}{c}{measurement}           \\
\cmidrule(lr){2-3} 
            & bound                        & free       \\
\midrule
bound       & deductive knowledge          & inductive bias (inference) \\
free        & deductive bias (consistency) & inductive knowledge \\
\bottomrule
\end{tabular}
\caption{Interpreting probing with different calibration and measurement datasets.}
\label{table:probing_interps}
\end{table}

\paragraph{Probing with free vs. bound splits.}
\Cref{table:probing_interps} details four possible probing setups when separating the auxiliary dataset $D$ into free and bound splits. In particular, when calibration and measurement occur on the same split, the probe quantifies the \textbf{knowledge}, or information content, that can be extracted from the LM representations; conversely, probing with different splits measures the transferability of the representations across different splits, which relies on implicit \textbf{bias}. Additionally, because the bound variables outcomes, can, by definition, be deduced from the given data (and hypothesis class $\mathcal{M}$), measuring on the bound split relates to the \textbf{deductive} ability of the LM; conversely, measuring on the free split is inherently an \textbf{inductive} process. We highlight that the inductive bias can be understood as quantifying the capacity of the LM representations to \emph{infer values in unseen data by applying theories derived from known data}, a form of inductive inference; while the deductive bias measures the extent to which the LM representations \emph{produce theories of unseen data that are consistent with the observed data}, a key tenet of deductive logic.

\subsection{Causal mediation analysis of probing}

\begin{figure}
\begin{center}
\includegraphics[scale=.85]{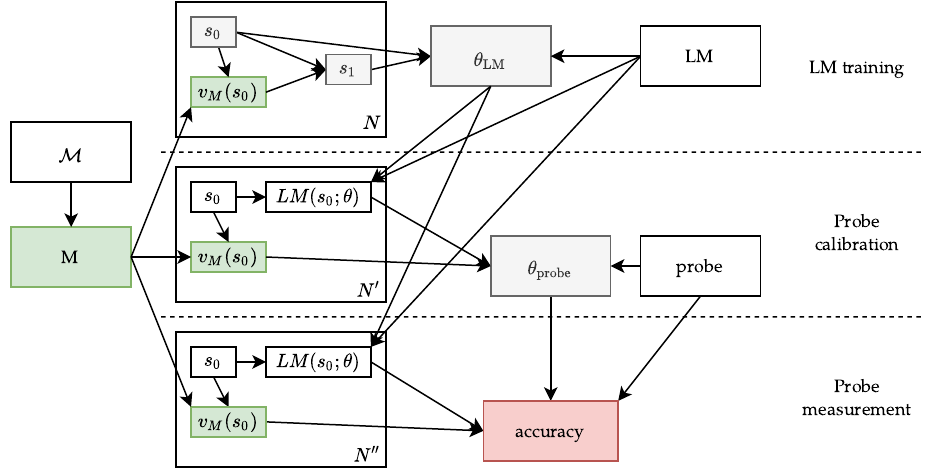}
\end{center}
\caption{An SCM depicting the LM training, probe calibration, and probe measurement. We use plate notation for repeated iid samples, e.g., we draw $N$ samples for LM training.}
\label{fig:full_SCM}
\end{figure}

We next turn to controlling for the probe (\textbf{Q1}). 
Intuitively, the challenge is any measurement using a supervised probe conflates the LM's representation of the auxiliary labels with the probe's ability to learn the auxiliary task \citep{hewitt2019designing}. While there exist a number of proposals for controlling for the contribution of the probe, such techniques typically do not provide any formal guarantees, rendering their correct application and interpretation a challenge \citep{belinkov-2022-probing}.

We propose a method of disentangling the two effects using the formal framework of causal mediation analysis, and specifically, path-specific effects, which analyze how causal effects decompose over multiple causal paths \citep{avin2005identifiability,imai2010general}. To begin, we extend the SCM of the data generation process to include (1) the LM training, (2) the probe calibration, and (3) the probe measurement. \Cref{fig:full_SCM} illustrates an example where the hypothesis class $\mathcal{M}$ consists of 3-variable SCMs with a single exogenous, observed variable $s_0$, a latent variable $v_M$, and another observed variable $s_1$.

Observe that there are three causal paths from the true SCM of the data generation process to the auxiliary task accuracy, each of which is mediated by a different set of the latent variables $v_M$: (1) during LM training, the LM is trained on a dataset whose text is causally affected by $v_M$; (2) during probe calibration, the probe is calibrated using $v_M$ directly; and (3) during probe measurement, the probe is evaluated for accuracy on $v_M$ directly. However, we only care to measure the effect over the first of these causal paths, i.e.:
\begin{center}
    \emph{To what extent can the auxiliary task performance be attributed to what LM learns from the latent variables in its training data?}
\end{center}
This question can be posed formally as the necessary indirect effect of the paths mediated by the LM's learned representation for some baseline causal dynamics $M'$:
\begin{multline*}
    \text{NIE}_{M, M'}(\theta_\text{LM}) = \\
    \mathbb{E}\big[\text{accuracy}
    \mid \text{LM is trained on $M$, probe is calibrated and measured on $M$}\big] \\ - \mathbb{E}\big[\text{accuracy}
    \mid do\text{(LM is trained on $M'$), probe is calibrated and measured on $M$}],
\end{multline*}

Although path-specific effects offer a crisp conceptual framework for isolating the contribution of the LM in probing experiments, actually computing NIE is not straightforward. First, picking a proper baseline $M'$ is critical; intuitively, if we pick an inappropriate $M'$, then the NIE will measure the difference between $M$ and $M'$ in addition to the latent variables hypothetically mediated by the LM representations. Second, measuring the effect requires intervening along the path of interest while holding the other paths constant, i.e., we would need to retrain LM according to the baseline $M'$, which would render the technique prohibitively expensive for large pretrained LMs. 

Let $acc_{aux}(M_0, M_1)$ denote the (expected) auxiliary task accuracy after the LM is trained using the SCM $M_0$ and the probe is calibrated and measured on $M_1$. The following result addresses these challenges (proof in \Cref{app:proofs}).

\begin{definition}[Valid baseline]
\label{def:valid}
$M'$ is a \textbf{valid baseline} for $M$ if
\begin{align}
\label{eq:1}
acc_{aux}(M', M') &\ge acc_{aux}(M, M) \\
\label{eq:2}
acc_{aux}(M, M') &\ge acc_{aux}(M', M).
\end{align}
\end{definition}

\begin{proposition}
\label{prop:valid}
Let $M'$ be a \textbf{valid baseline} for $M$. Then
$$acc_{aux}(M, M) - acc_{aux}(M, M') > 0$$
implies both $\text{NIE}_{M, M'}(\theta_\text{LM})>0$ \emph{and} $\text{NIE}_{M', M}(\theta'_\text{LM})>0$. 
\end{proposition}

Intuitively, $M'$ is a valid baseline when measuring $M'$ is easier than measuring $M$ under both normal or intervened circumstances. The conclusion then states that, so long as $acc_{aux}(M, M) - acc_{aux}(M, M') > 0$ (which can be evaluated by running probe calibration and measurement twice rather than training the LM twice), there is no bias in which SCM is used to train the LM and which is the baseline: the LM representations always mediate a positive amount of the measured effect.

A positive NIE now also has a rigorous interpretation as the LM having \emph{learned latent concepts}, as some positive amount of causal effect is transferred through the representations of the LM. For instance, a positive mediated measurement for inductive bias implies that 
\begin{center}
    \emph{The presence of latent causal variables in the pretraining data causes the LM to learn representations that generalize to unknown data.}
\end{center}

\subsection{Discussion}
\label{sec:steps}

We summarize the \textbf{latent causal probing} framework as follows:
\begin{enumerate}
\item Select a set of exogenous, latent, and observed variables and pick a hypothesis class $\mathcal{M}$ of SCMs (from the set of all Markovian SCMs over the variables).
\item Fix a specific target SCM $M \in \mathcal{M}$ and a set of latent variables $v \in M$ to test.
\item Construct the auxiliary dataset and create the bound vs. free partition (if possible).
\item Identify a valid baseline $M'$ and perform the mediation analysis.
\end{enumerate}
A significant measurement $acc_{aux}(M, M) - acc_{aux}(M, M') > 0$ is interpreted evidence that the LM encodes the latent concepts in its representations. 
We conclude with some remarks.

\paragraph{Interventions, and probing for non-causal latent variables.}
Our mediation technique requires knowing ``what would the text have been if the underlying dynamics were different?'', which could be difficult (especially in non-synthetic domains). Similarly, for non-causal latent variables, such as part-of-speech, producing a hypothesis class $\mathcal{M}$ with more than one SCM may not be possible: what would the data look like in a counterfactual world in which ``dog'' is actually an adverb? Unfortunately, our analysis suggests that a baseline $M'$ which induces a different distribution of text is a necessary precondition, since otherwise $\text{NIE}_{M, M'}(\theta_\text{LM})$ and $\text{NIE}_{M', M}(\theta'_\text{LM})$ cannot both be positive (as $M$ and $M'$ are indistinguishable when used to train the LM). Intuitively, we interpret this result as saying \emph{any measurement is inherently biased when the auxiliary task has only one ``right'' answer}.

\paragraph{Probe architecture and hyperparameters.}
Our framework also explicates the role of the probe's architecture and other hyperparameters in the training process, such as the optimizer, learning rate, dataset size, etc., as potential \emph{moderators}, but not mediators (\textbf{Q2}). In other words, so long as there exists a choice of hyperparameters such that the NIE is positive, the analysis concludes that there exists a causal effect mediated by the model's parameters (although certain settings of the moderator variables could offer additional interpretations). Practically speaking, our framework also offers a novel way to interpret (and justify) complex probes \citep{voita-titov-2020-information,pimentel-cotterell-2021-bayesian}.

\section{Experiments}

We conduct empirical study of whether an LM, trained from scratch on a corpus of program data, induces the latent concepts in the underlying data generation process.

\subsection{Methods}

We describe the key steps according to the framework in \Cref{sec:steps}; \Cref{app:experiment_details} contains full experimental details (e.g., LM and probe architecture and training, LM representations).

\paragraph{Hypothesis class.} The exogenous variables are the initial state and program. The latent variables are the intermediate states, and the observed variables are the initial and final state and the program. The hypothesis class $\mathcal{M}$ is all Markovian SCMs over the variables.

\paragraph{Target SCM and latent variables.} The target SCM $M \in \mathcal{M}$ is the true data generation process in \Cref{fig:exp_SCM}. 
The target latent variables consist of the robot's position, facing direction, and whether the robot is facing a rock for each intermediate state.

\paragraph{Auxiliary dataset construction and bound and free latent variable outcomes.} For the auxiliary dataset, we use the same data generation process, except that programs range in length between 1 and 15, and we replace the final state in the specification with the initial state.
Due to the size of the training corpus (1 million samples), we assume the LM observed all combinations of the exogenous variables. Because the programs in the LM training corpus are between length 5 and 9, the bound latent variables are $s_6$ to $s_{10}$ (they are observed during training as the final state). The free latent variables are $s_1$ to $s_5$ and $s_{11}$ to $s_{15}$.

\paragraph{Valid baseline.} We construct valid baselines according to a counterfactual state of the world where the intermediate states are generated by executing the program according to a different set of causal dynamics. Specifically, we define $M'$ using the same SCM, but permute the causal dynamics of the \texttt{turn\_right}, \texttt{turn\_left}, and \texttt{move} actions (e.g., the robot turns left when executing a \texttt{turn\_right} action). As $M$ and $M'$ are clearly symmetric from a language modeling perspective, \Cref{def:valid} (and hence \Cref{prop:valid}) holds.

\begin{figure}[htb]
 \centering
 \begin{subfigure}[b]{0.48\textwidth}
     \centering
     \includegraphics[width=.75\textwidth]{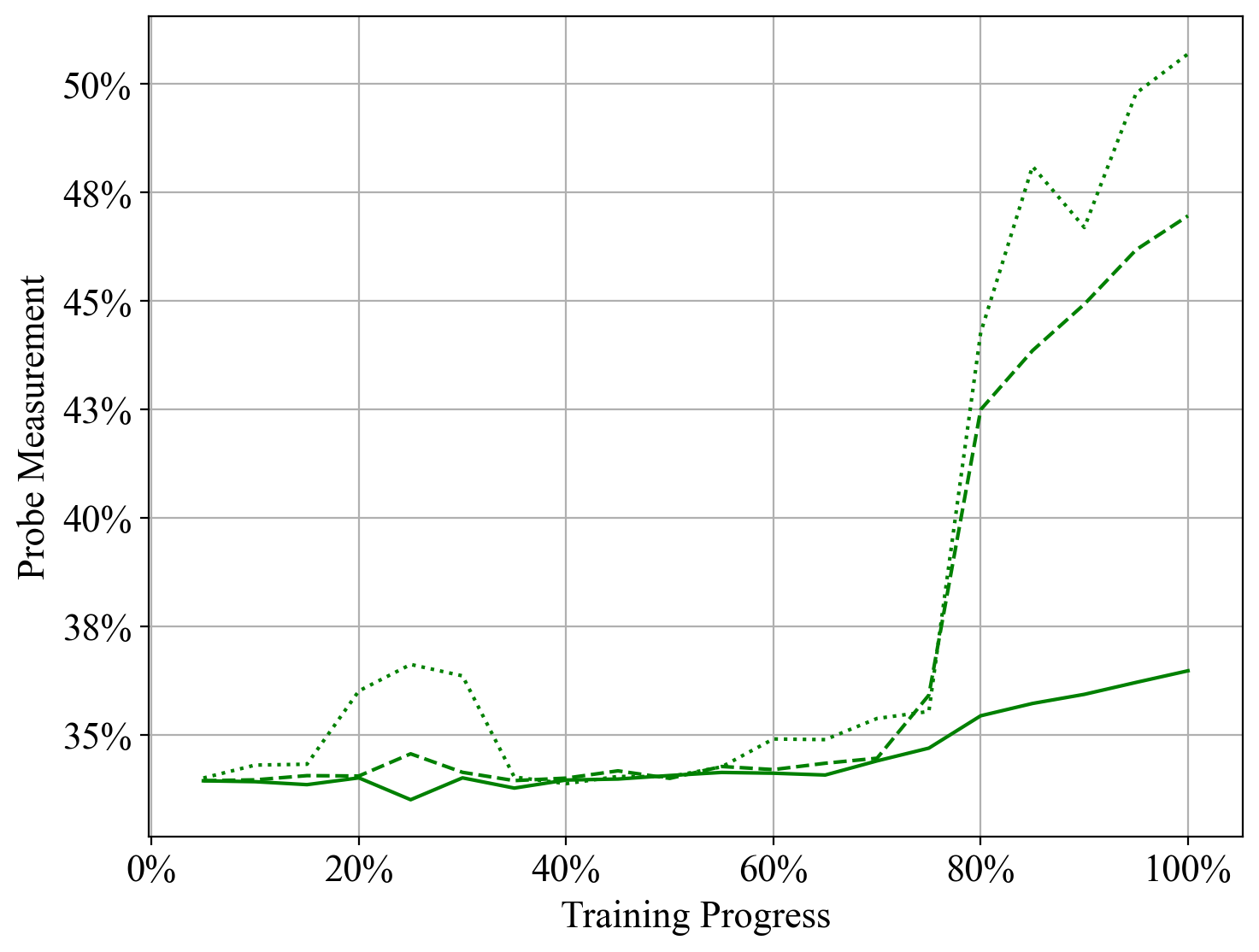}
     \caption{Deductive knowledge.}
     \label{fig:id}
 \end{subfigure}
 \hfill
 \begin{subfigure}[b]{0.48\textwidth}
     \centering
     \includegraphics[width=.75\textwidth]{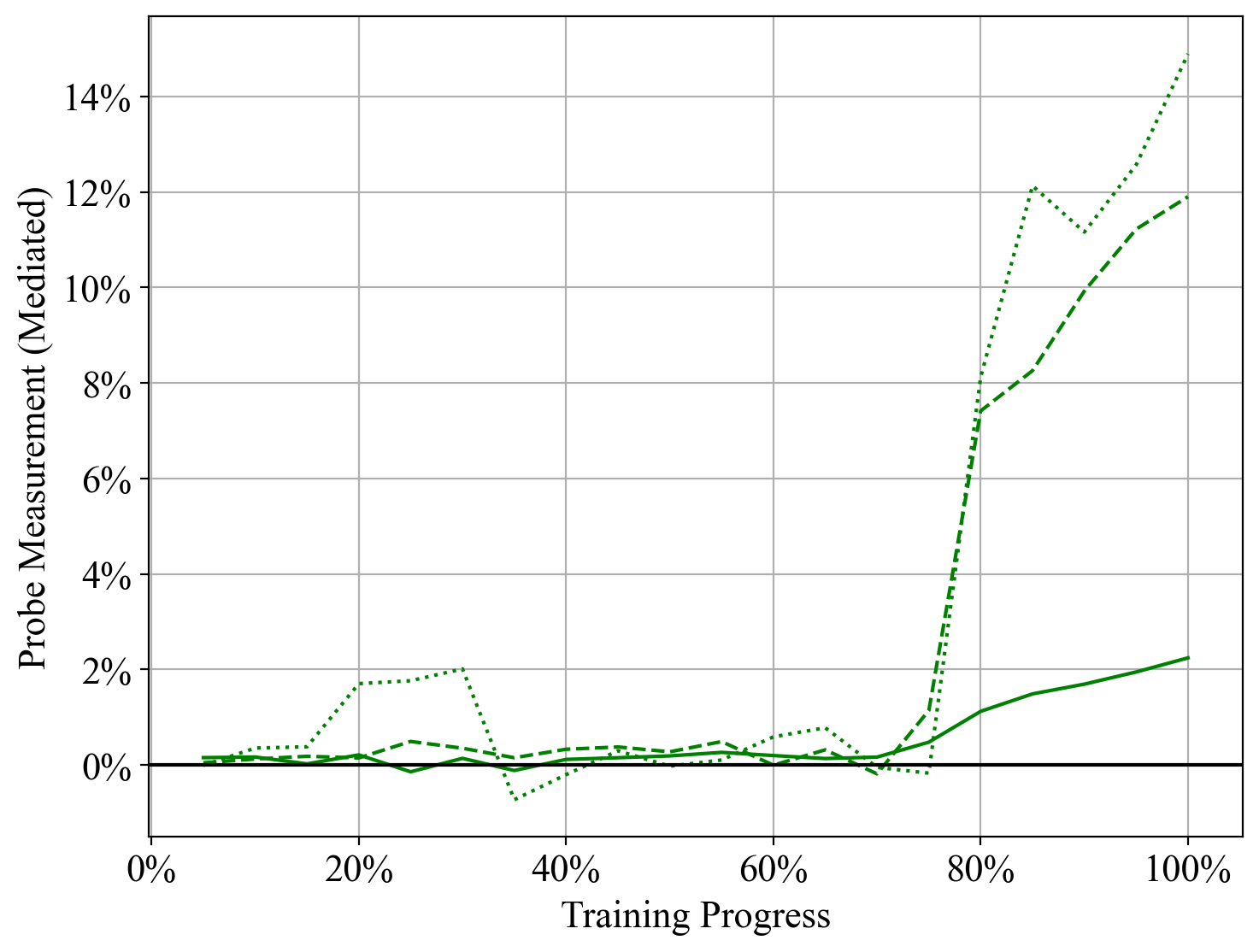}
     \caption{Deductive knowledge, mediated.}
     \label{fig:id:mediated}
 \end{subfigure} \\
 
 \begin{subfigure}[b]{0.48\textwidth}
     \centering
     \includegraphics[width=.75\textwidth]{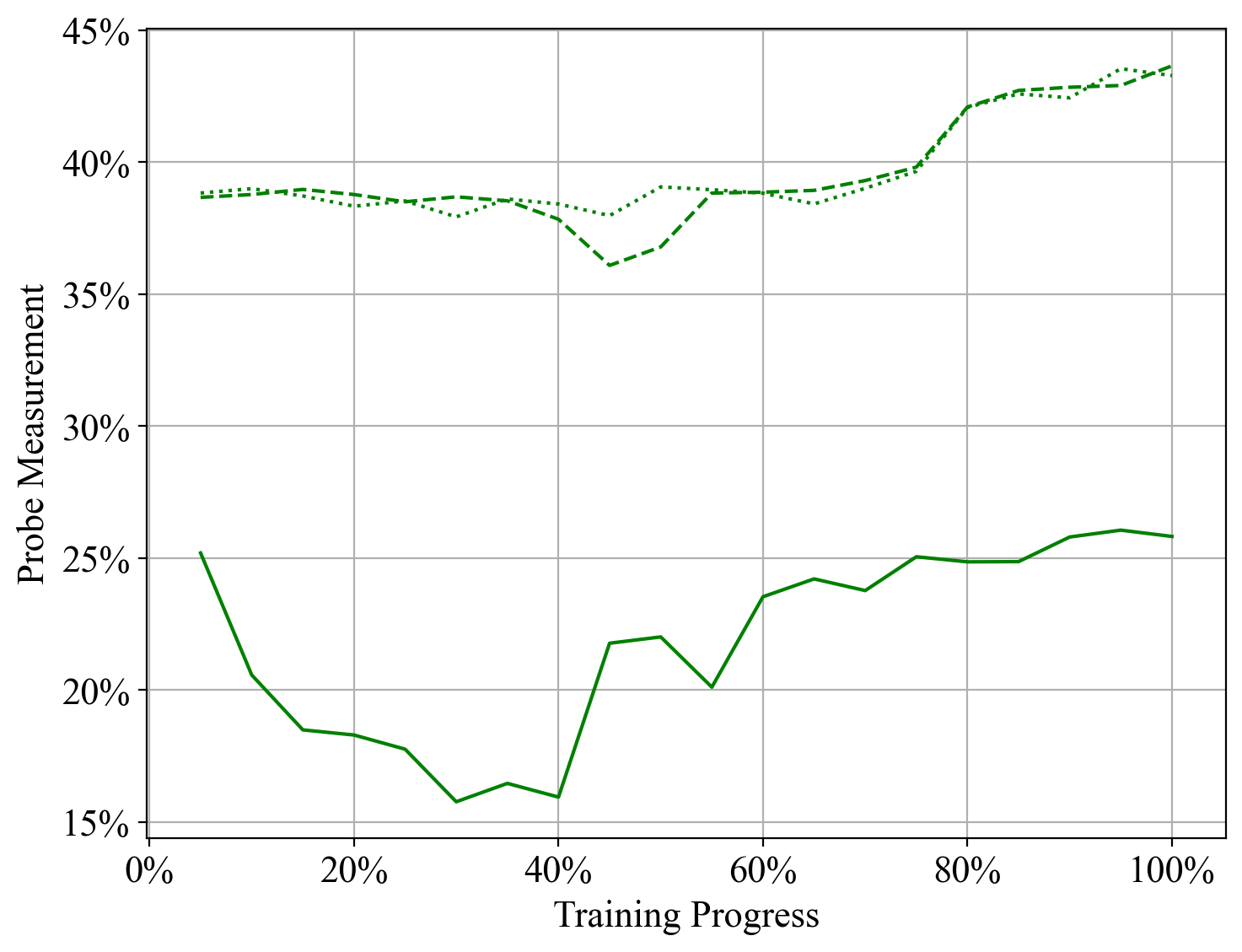}
     \caption{Inductive bias.}
     \label{fig:ig}
 \end{subfigure}
 \hfill
 \begin{subfigure}[b]{0.48\textwidth}
     \centering
     \includegraphics[width=.75\textwidth]{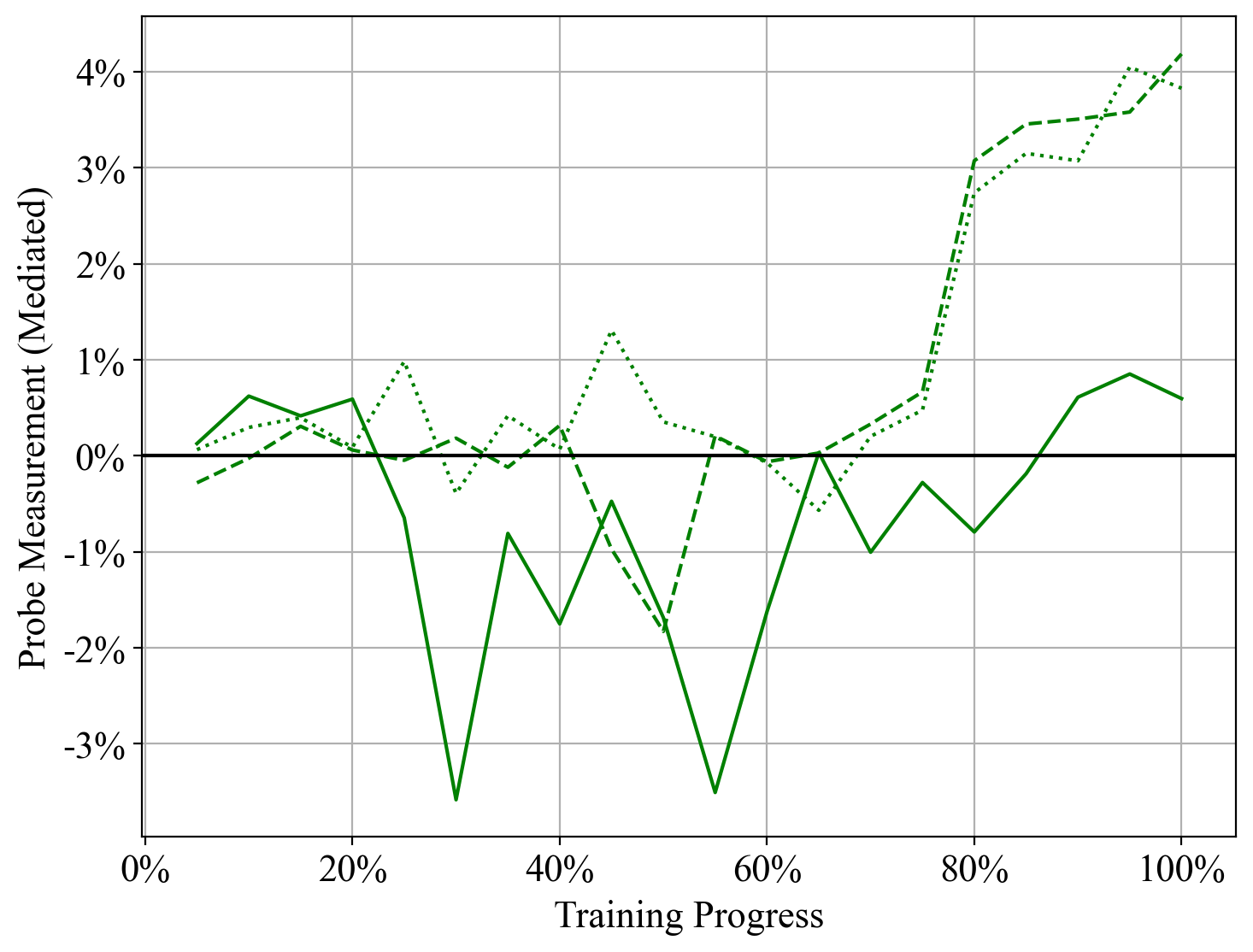}
     \caption{Inductive bias, mediated.}
     \label{fig:ig:mediated}
 \end{subfigure} \\
 
 \begin{subfigure}[b]{0.48\textwidth}
     \centering
     \includegraphics[width=.75\textwidth]{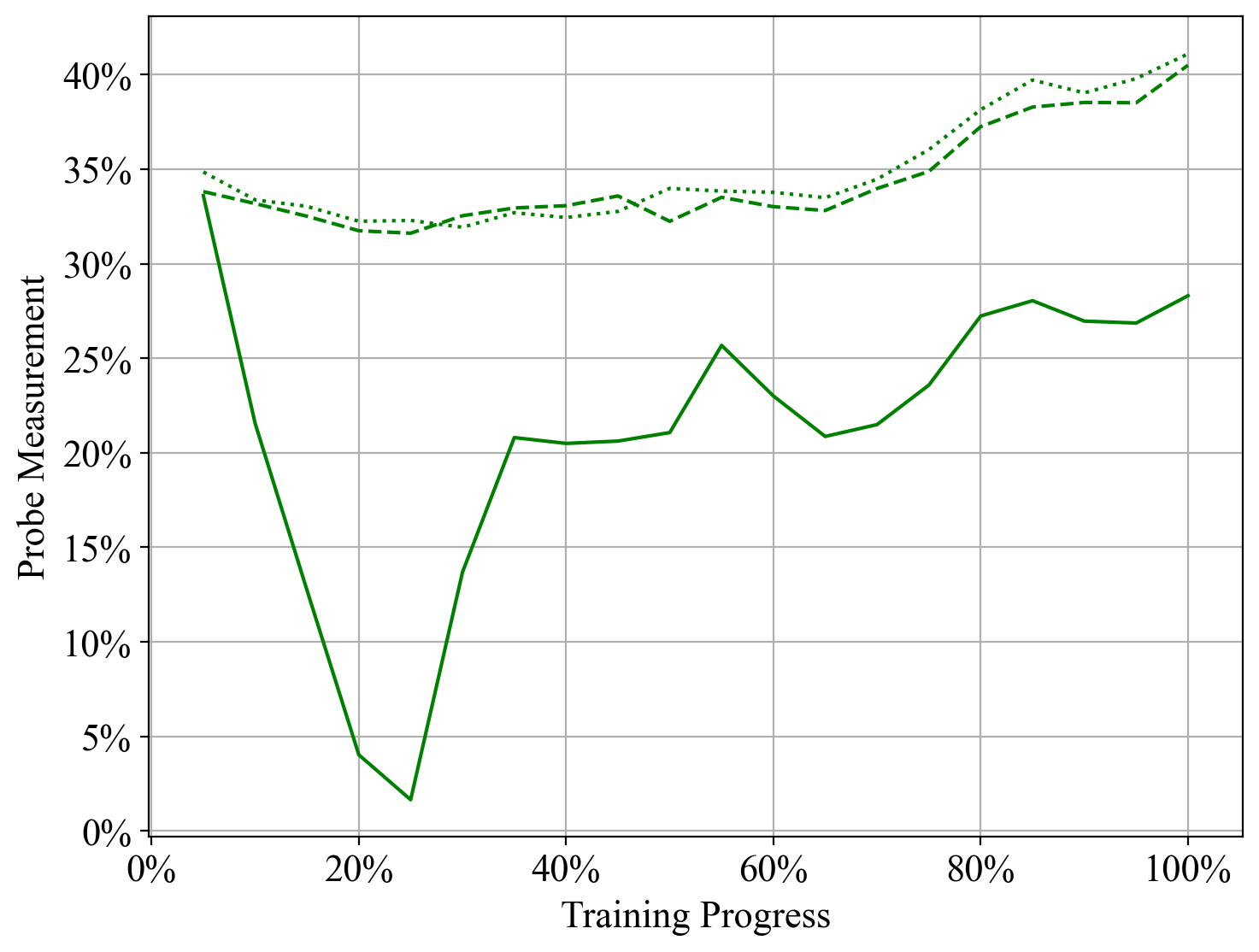}
     \caption{Deductive bias.}
     \label{fig:ia}
 \end{subfigure}
 \hfill
 \begin{subfigure}[b]{0.48\textwidth}
     \centering
     \includegraphics[width=.75\textwidth]{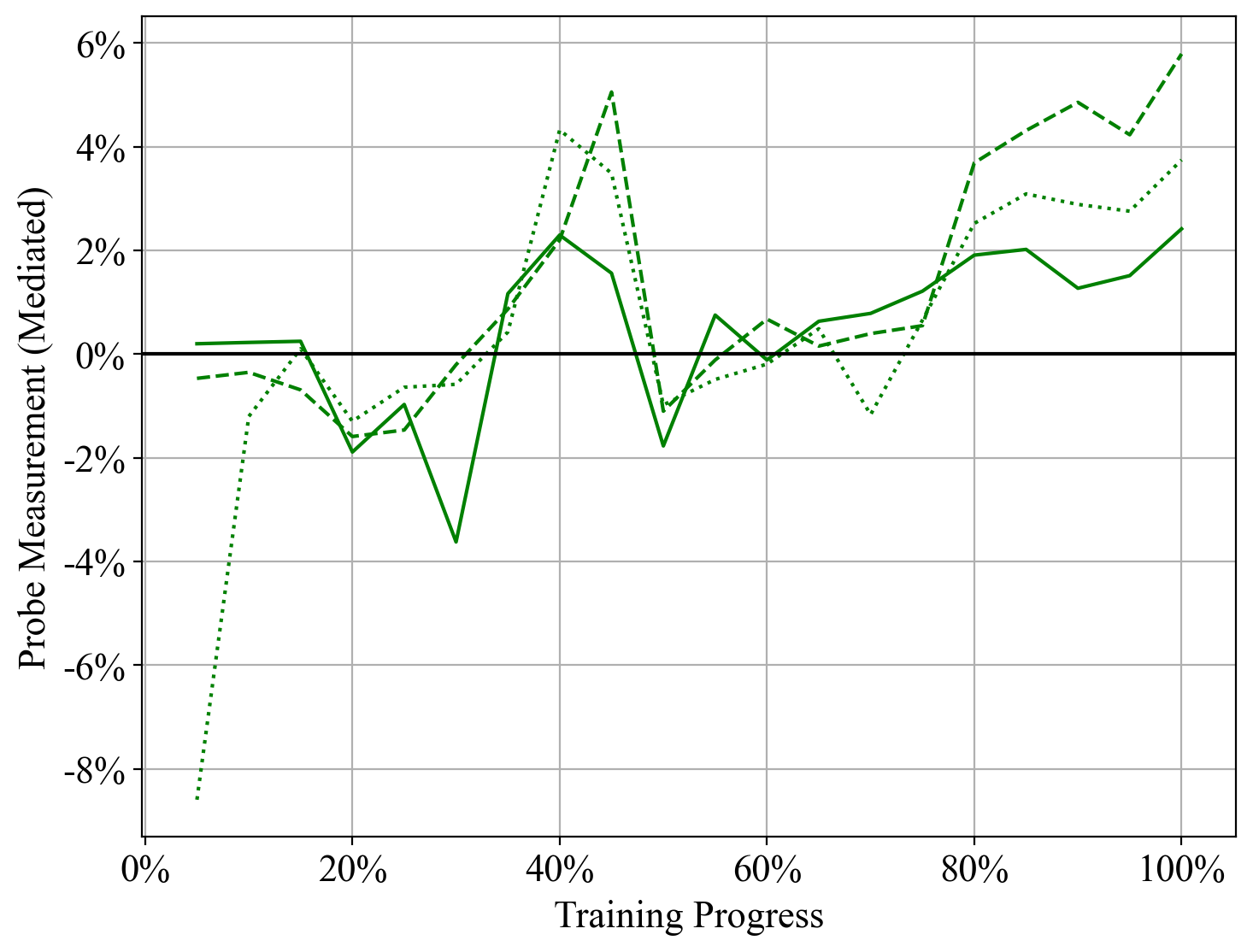}
     \caption{Deductive bias, mediated.}
     \label{fig:ia:mediated}
 \end{subfigure} \\
 
 \begin{subfigure}[b]{0.48\textwidth}
     \centering
     \includegraphics[width=.75\textwidth]{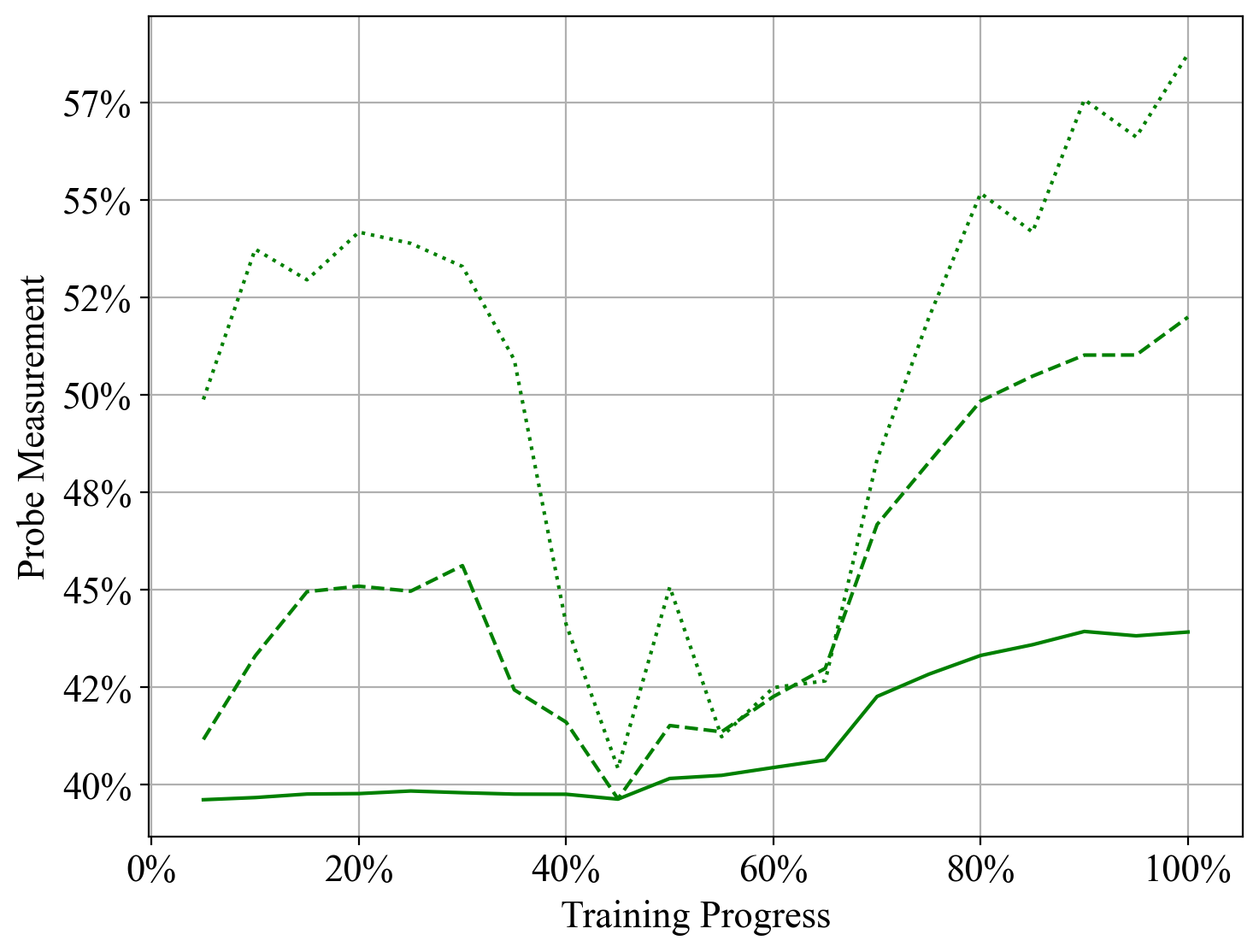}
     \caption{Inductive knowledge.}
     \label{fig:ood}
 \end{subfigure}
 \hfill
 \begin{subfigure}[b]{0.48\textwidth}
     \centering
     \includegraphics[width=.75\textwidth]{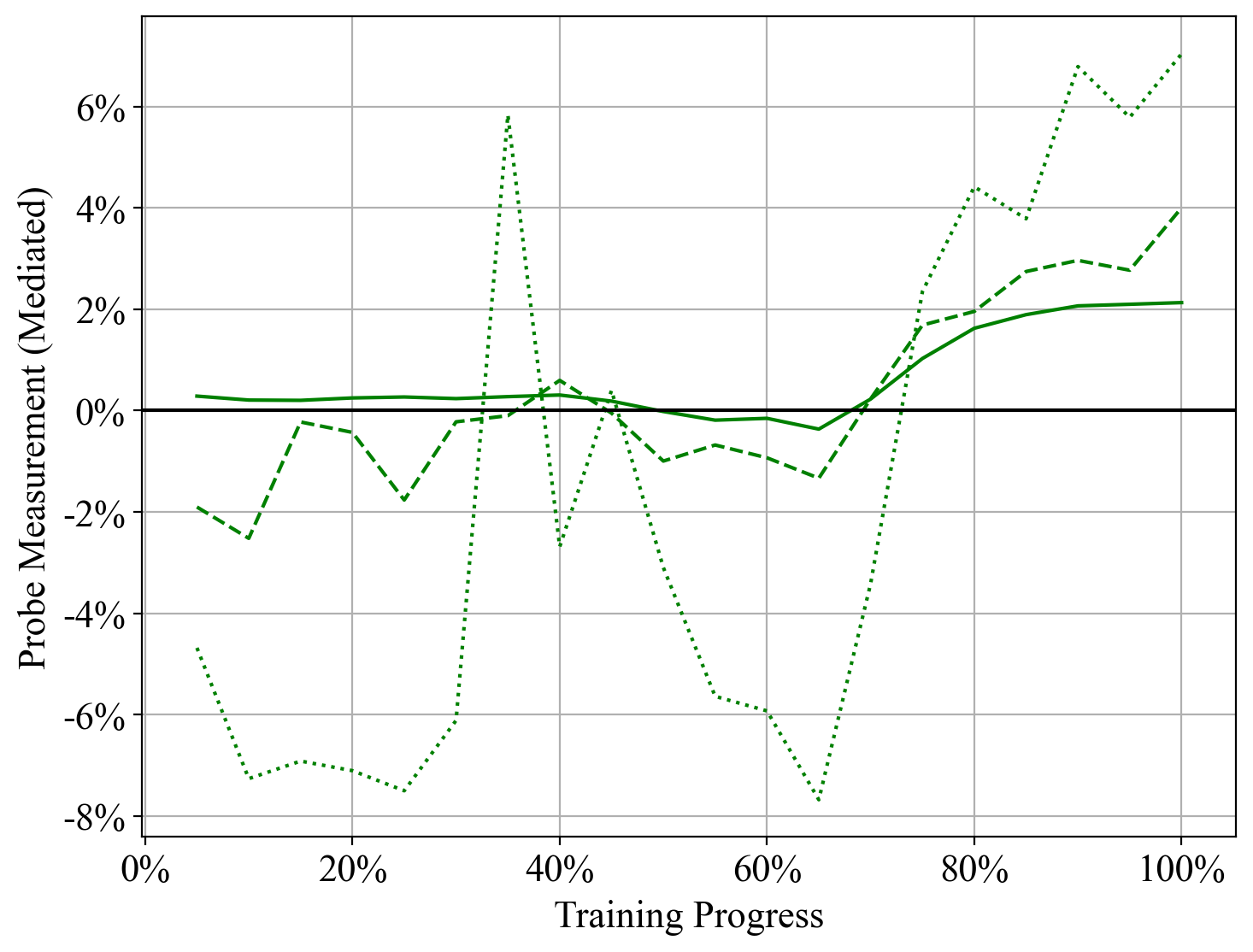}
     \caption{Inductive knowledge, mediated.}
     \label{fig:ood:mediated}
 \end{subfigure}
 
\caption{Results from the main experiments. Solid, dashed, and dotted green lines plot the accuracy of a linear, 1-layer MLP, and 2-layer MLP probing classifier, respectively. The final model achieves 92.4\% accuracy on generating correct programs for unseen specifications.}
\label{fig:results}
\end{figure}

\subsection{Results}

\Cref{fig:results} plots the main results. For all four measurements (deductive knowledge, inductive bias, deductive bias, and inductive knowledge) and across all three probes (linear, 1-layer MLP, 2-layer MLP), the mediated measurements are significantly positive (above the dashed line at 0\%) by the end of training. We thus conclude that a positive fraction of the \emph{observed measurements of latent concepts can be attributed to what is learned by the LM's representations}.
We make several additional observations:

\paragraph{Deeper probes are more (generally) more accurate.} The linear probe exhibits the lowest raw and mediated measurements of the 3 probes across all four tasks. Furthermore, the 2-layer MLP exhibits the highest raw measurement in all four tasks, and the highest mediated measurement in two of the four tasks; and only in the case of the deductive bias (\Cref{fig:ia:mediated}) does the 1-layer MLP achieve a substantially higher measurement.
This suggests that more complex auxiliary tasks require more complex probes and, more generally, highlights the importance of probing frameworks that can robustly accommodate deeper probes. 

\paragraph{Raw measurements can be uncorrelated with mediated measurements.}
Comparing deductive and inductive knowledge, the raw measurement for inductive knowledge is the substantially higher (inductive knowledge: $\sim$52\% and $\sim$58\% vs. deductive knowledge: $\sim$48\% and $\sim$51\%, using a 1-layer and 2-layer MLP, respectively, at the end of training) but the mediated measurements are reversed (inductive knowledge: $\sim$4\% and $\sim$7\% vs. deductive knowledge: $\sim$12\% and $\sim$15\%, using a 1-layer and 2-layer MLP, respectively, at the end of training). These results confirm that the raw measurements of probing experiments may be misleading in that a higher raw measurement does not necessarily mean that the LM performs the auxiliary task better.

\Cref{app:experiment_results} contains further results and analyses.

\section{Related work}

\paragraph{Causal interpretability of LMs.}
Several prior lines of work apply causal techniques to the interpretability of LMs. These works typically intervene on either the model's representations \citep{2021amnesiac,geiger2021causal,meng2022locating,abraham2022cebab,li2022emergent}
or the model's inputs \citep{Kaushik2020Learning,vig2020investigating,gangal-hovy-2020-bertering,amini2023}, and analyze the causal effect on the LM's outputs.
In contrast, we present a formal framework that, conceptually, intervenes on the model's \emph{training data}, and measures the causal effect on the LM's internal representations. To the best of our knowledge, \citet{elazar2022measuring} is the only other work that studies the causal relationship between the training data and the LM, presenting a technique for estimating the causal effect of dataset statistics on the factuality of LM's outputs.

\paragraph{World models in LMs.}
The hypothesis that LMs can induce latent concepts is related to evidence of world models in LMs, in which the agenda is to understand the extent to which LMs are capable of grounding their inputs to (some representation of) reality. For instance, 
\citet{abdou-etal-2021-language} show that LMs' representations of color terms have geometric structure that reflects human perceptual distance.
\citet{li-etal-2021-implicit} find that LMs perform entity state tracking over the course of simple stories.
\citet{li2022emergent} present evidence than an LM trained on Othello transcripts develops a representation of the underlying board state.
Our work is based on \citet{jin2023evidence}, who show that an LM trained only on sequences of instructions and instances of initial and final states in a grid-world navigation task can develop representations of the intermediate states resulting from the sequence of instructions is applied. All these works use some form of supervised classifier trained for an auxiliary objective, but unlike the present work, do not provide any rigorous guarantees that control for the classifier's ability to fit the auxiliary objective independently of whether the LM has truly learned a world model (or simply represents the text as is). The closest is \citet{jin2023evidence}, whose \emph{semantic probing interventions} follow a similar interventional intuition. However, they do not present any formal analysis, and their baselines also fail to satisfy the requirements of a valid baseline; \Cref{appendix:comp_with_icml} discusses some of the differences in experimental design in greater detail.
Finally, our formulation of world models according to the underlying causal mechanism that generated the data is also highly related to the position developed by \citet{andreas-2022-language}, who argues that LMs could model properties of agents that are likely to have uttered the language in their training data.

\paragraph{Frameworks for probing.} A number of works have proposed frameworks toward a more rigorous understanding of probing. One line takes an information-theoretic view on the information represented by the LM \citep{zhu2020information,pimentel2020information,voita-titov-2020-information,pimentel-cotterell-2021-bayesian}. Our work uses probes to quantify knowledge and biases in the LM's learned representations.
\citet{immer2022probing} propose an interpretation of probing as quantifying the inductive bias of pretrained representations for downstream tasks, but their framework differs significantly from ours in that the model is understood as a representation-probe pair. In contrast, our approach unifies the LM training and probe calibration procedures under a single causal framework for analysis, yielding formal guarantees for the control of probes. Our analysis also reveals settings in which prior techniques, such as control tasks \citep{hewitt2019designing}, can yield biased estimates of the intended auxiliary measurement. 

\section{Conclusion}

This paper presents \textbf{latent causal probing}, a probing framework that studies whether LMs induce latent variables as a byproduct of the language modeling objective. Our framework offers robust tools for interpreting experiment results through the lens of causal analysis, and in particular, rigorously controls for the probe's contribution in learning the auxiliary task. Experimentally, we extend a previous study of whether Transformers can infer the intermediate states that underlie a sequence of actions. Our results provide strong empirical evidence that LMs can induce latent concepts from textual pretraining.

\bibliography{colm2024_conference}
\bibliographystyle{colm2024_conference}

\appendix
\section{Additional experimental details and results}
\subsection{Language model and probe details}
\label{app:experiment_details}

Following \citet{jin2023evidence}, the language model is a 350M parameter CodeGen model \citep{nijkamp2022codegen} taken from the HuggingFace Transformers library \citep{huggingface}. The model was trained for 2.5 billion tokens, which was roughly 6 passes or 80000 training batches over the training corpus. We refer to \citet{jin2023evidence} for further details.

We next describe the design and training of the probing classifiers; these notes apply to all the probing experiments, unless otherwise noted. The linear probe is a single linear layer. The MLP probes have ReLU, batch\_norm, then dropout(p=.2) after each linear layer. The hidden dimensions of the 1-layer and 2-layer MLP probes were (256,) and (256, 1024), respectively. The auxiliary datasets consisted of 500000 randomly selected samples. To extract representations from the LM, we use the same strategy as \citet{jin2023evidence}, averaging the LM hidden states over the layer dimension after processing each program token. Probes were trained using AdamW  \citep{loshchilov2018decoupled} with weight decay of 1e-4. The learning rate starts at 0.01, then decays by .1 at 75\% and 90\% through training. All probes are trained for 2000000 steps using a batch size of 256.

For the mediated results reported in \Cref{fig:results}, we generated the auxiliary dataset using an SCM that maps \texttt{turn\_right} to \texttt{turn\_left}, \texttt{turn\_left} to \texttt{move}, and \texttt{move} to \texttt{turn\_right}.

\subsection{Ablation studies}
\label{app:experiment_results}

This section present some ablation studies on the set up of the probing experiments.

\subsubsection{Valid baseline selection}

\begin{figure}[h!]
 \centering
 \begin{subfigure}[b]{0.48\textwidth}
     \centering
     \includegraphics[width=\textwidth,]{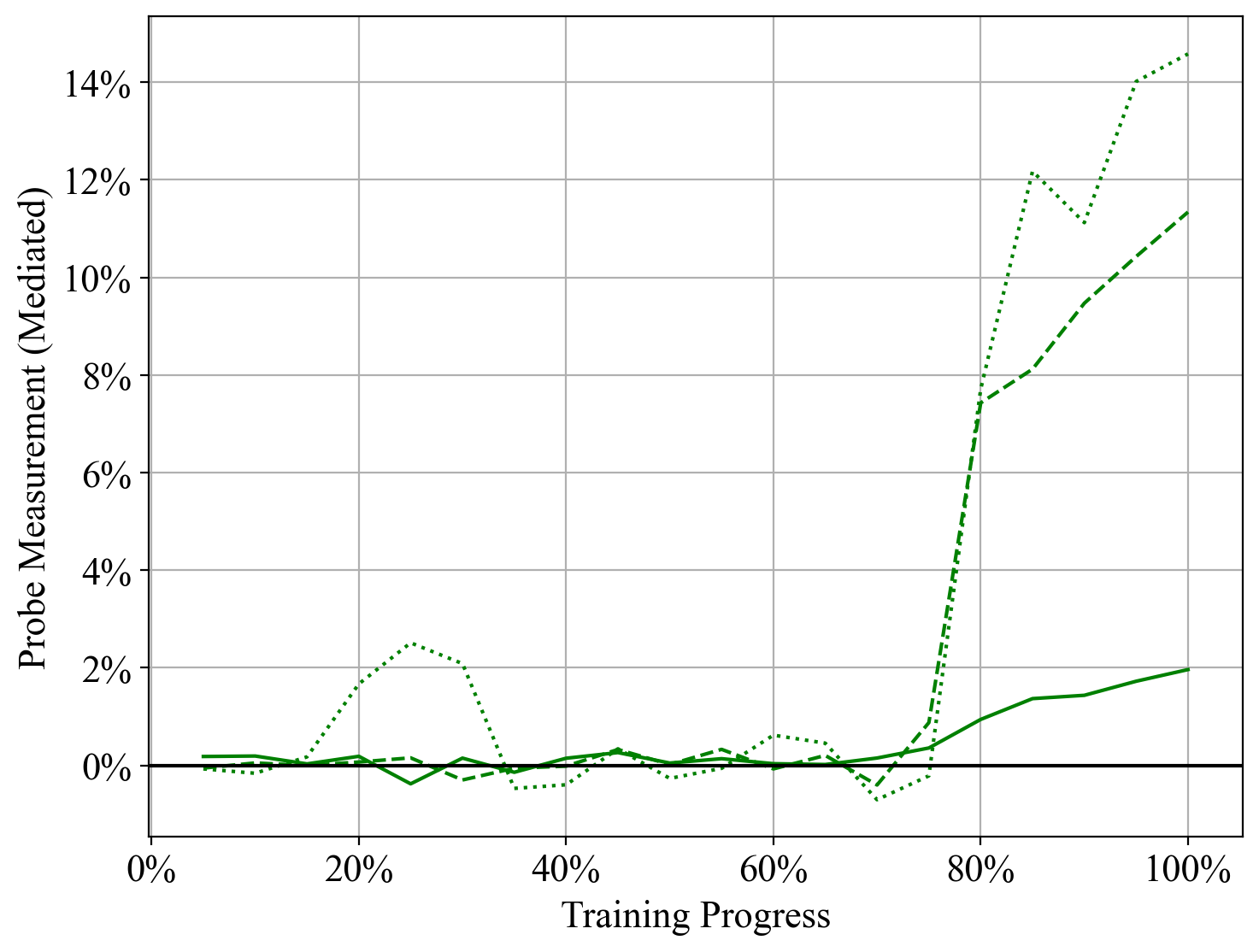}
     \caption{Deductive knowledge, mediated.}
     \label{fig:id:mediated8}
 \end{subfigure}
 \hfill
 \begin{subfigure}[b]{0.48\textwidth}
     \centering
     \includegraphics[width=\textwidth]{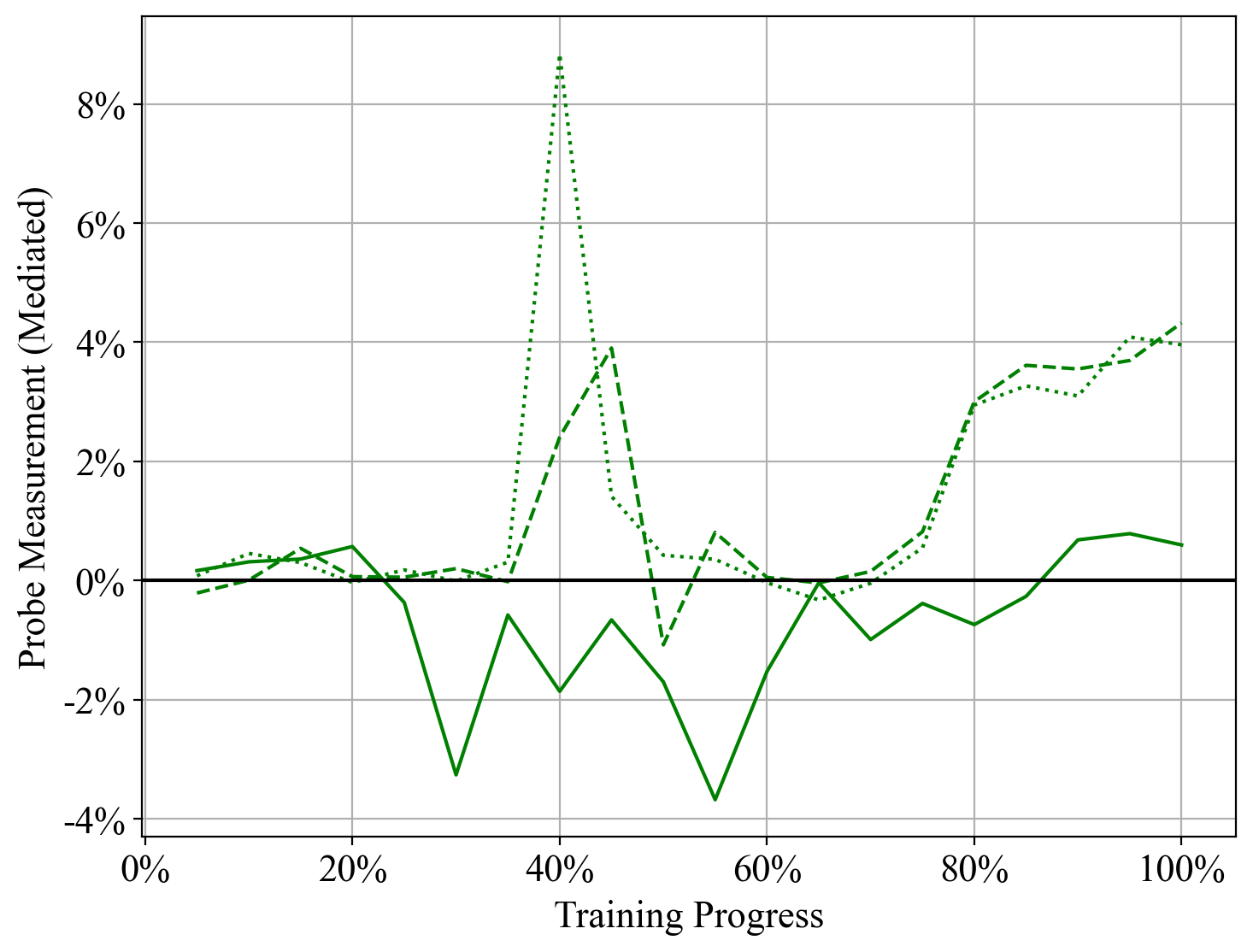}
     \caption{Inductive bias, mediated.}
     \label{fig:ig:mediated8}
 \end{subfigure}
 \\
 \begin{subfigure}[b]{0.48\textwidth}
     \centering
     \includegraphics[width=\textwidth]{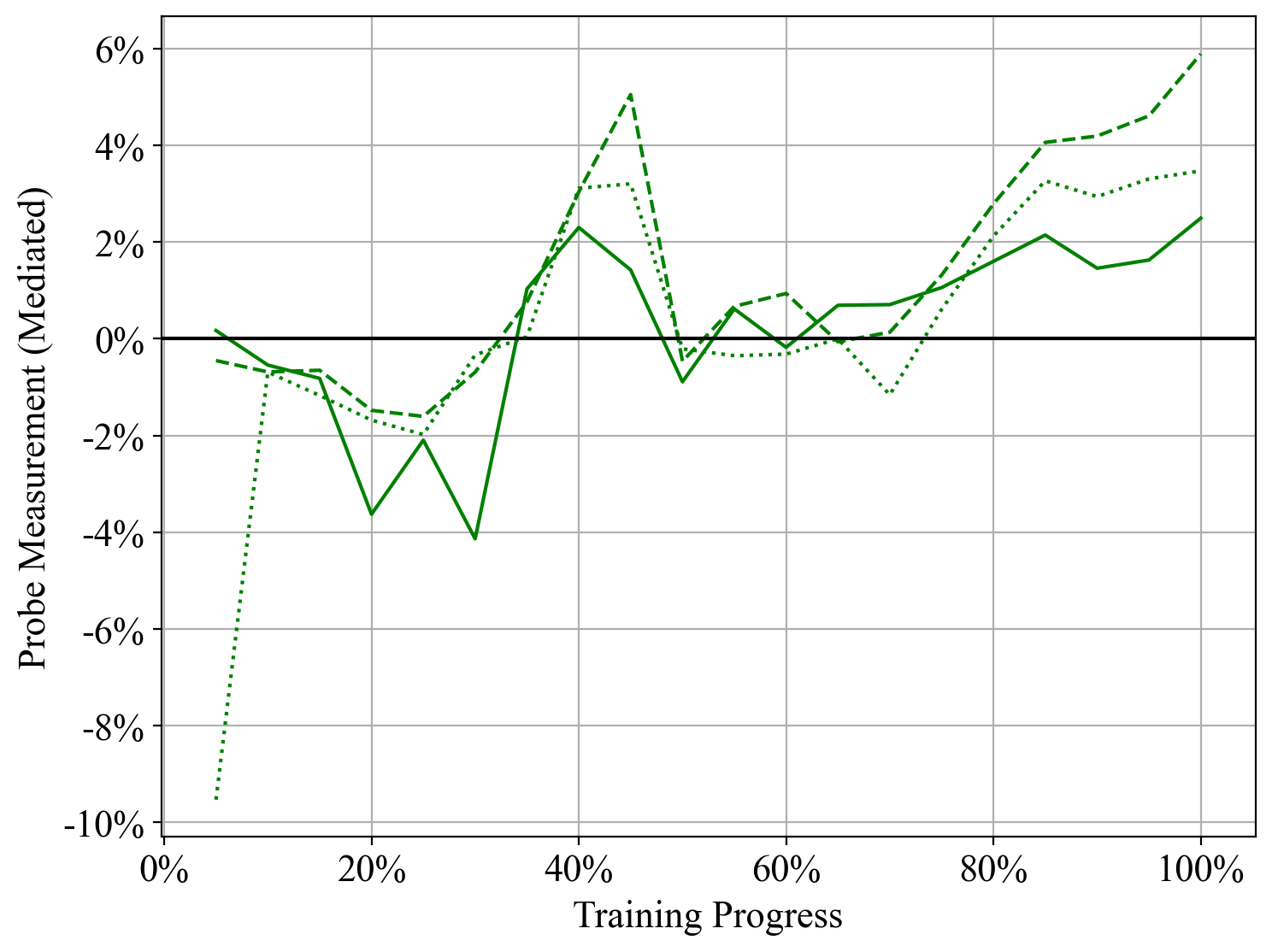}
     \caption{Deductive bias, mediated.}
     \label{fig:ia:mediated8}
 \end{subfigure}
 \hfill
 \begin{subfigure}[b]{0.48\textwidth}
     \centering
     \includegraphics[width=\textwidth]{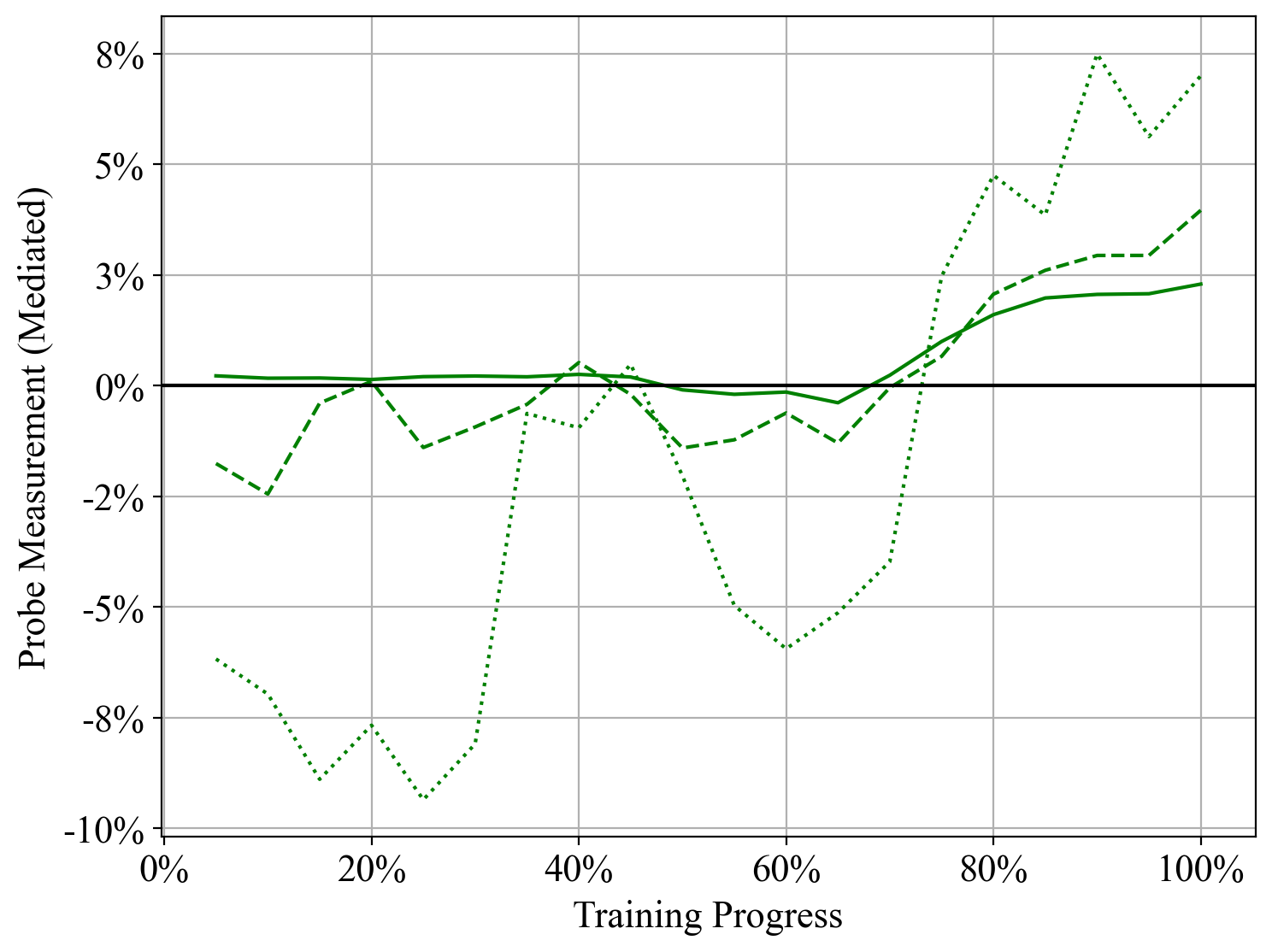}
     \caption{Inductive knowledge, mediated.}
     \label{fig:ood:mediated8}
 \end{subfigure}
\caption{Mediating with the valid baseline that swaps \texttt{move} and \texttt{turn\_left}.}
\label{fig:results_8}
\end{figure}

\begin{figure}[h!]
 \centering
 \begin{subfigure}[b]{0.48\textwidth}
     \centering
     \includegraphics[width=\textwidth,]{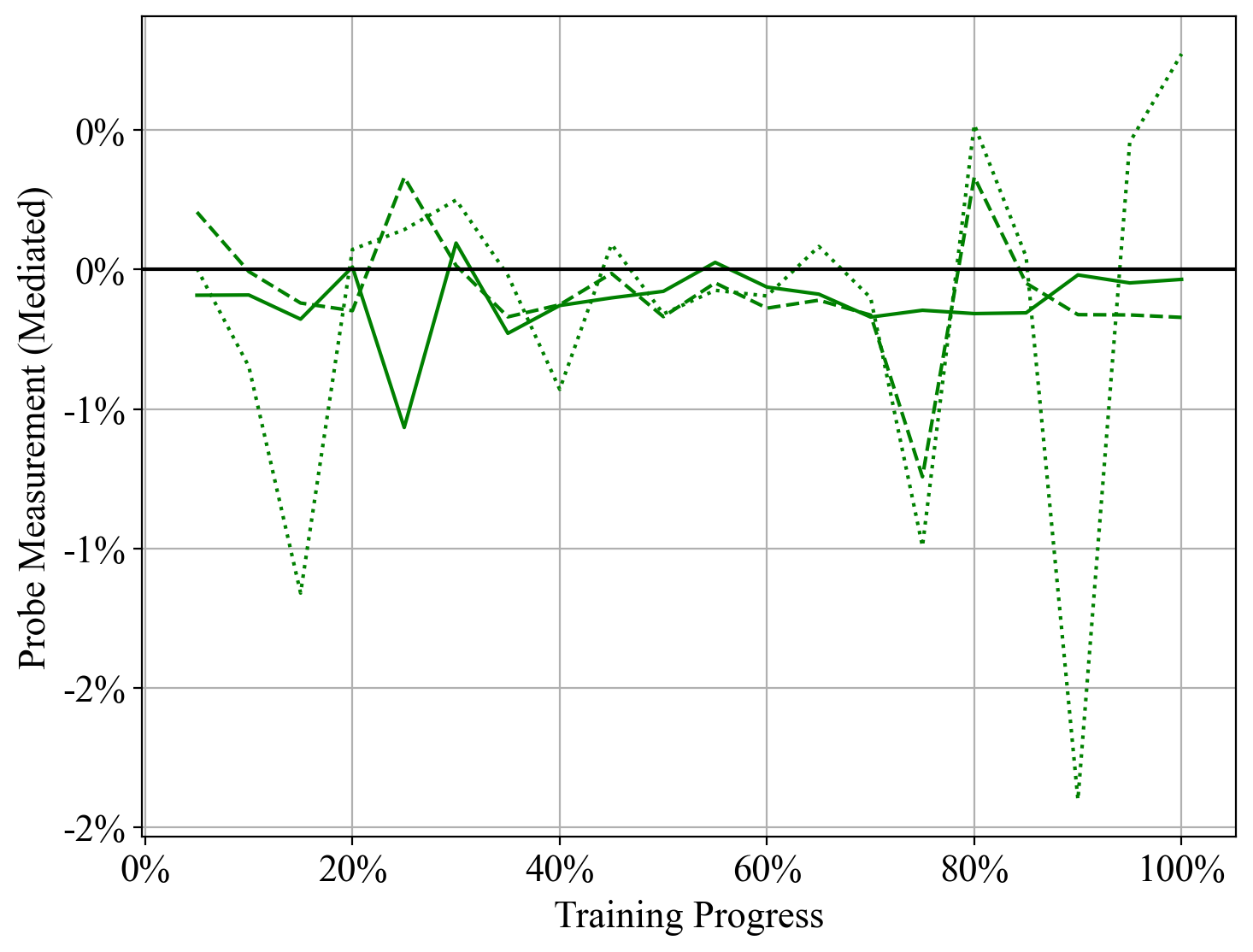}
     \caption{Deductive knowledge, mediated.}
     \label{fig:id:mediated7}
 \end{subfigure}
 \hfill
 \begin{subfigure}[b]{0.48\textwidth}
     \centering
     \includegraphics[width=\textwidth]{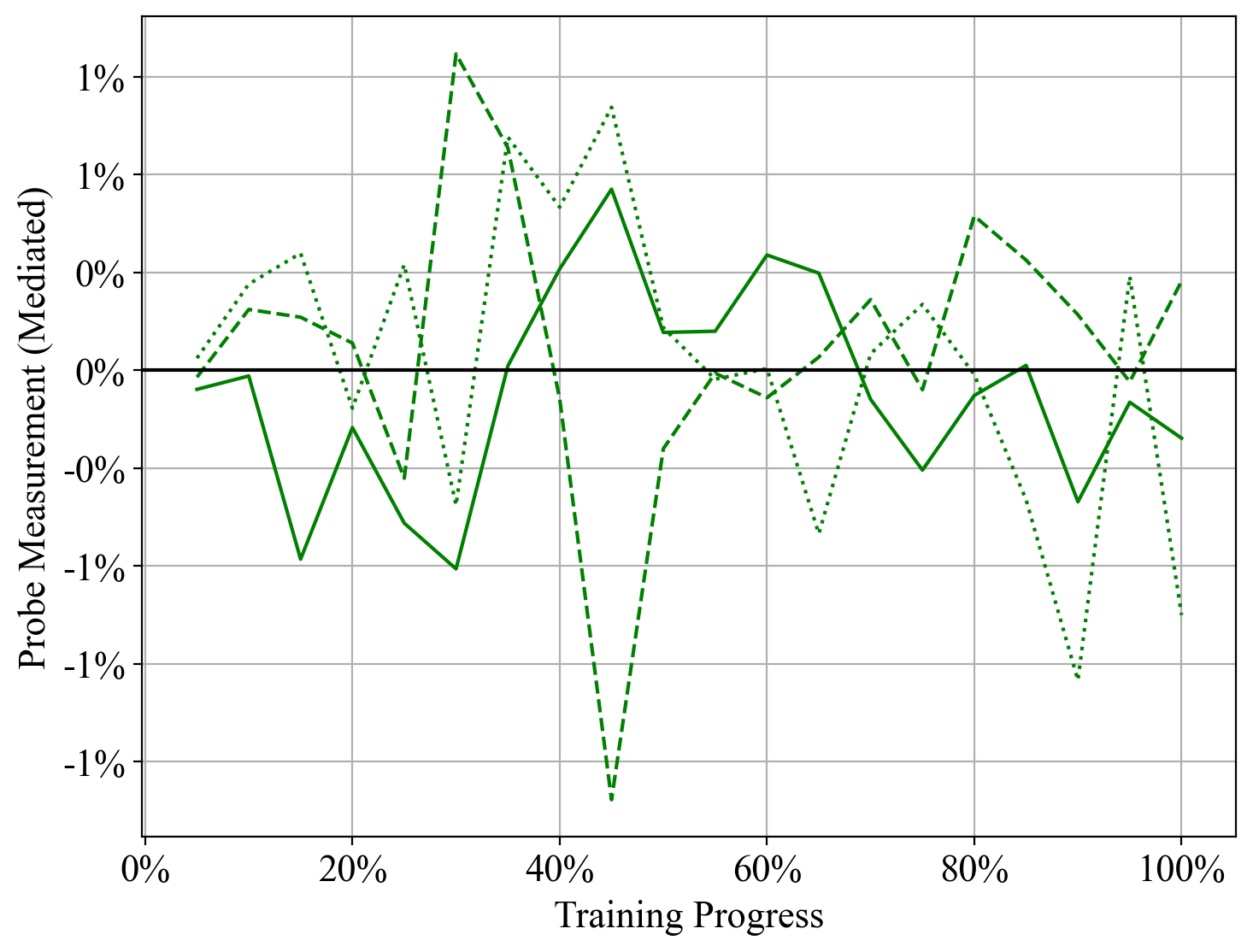}
     \caption{Inductive bias, mediated.}
     \label{fig:ig:mediated7}
 \end{subfigure}
 \\
 \begin{subfigure}[b]{0.48\textwidth}
     \centering
     \includegraphics[width=\textwidth]{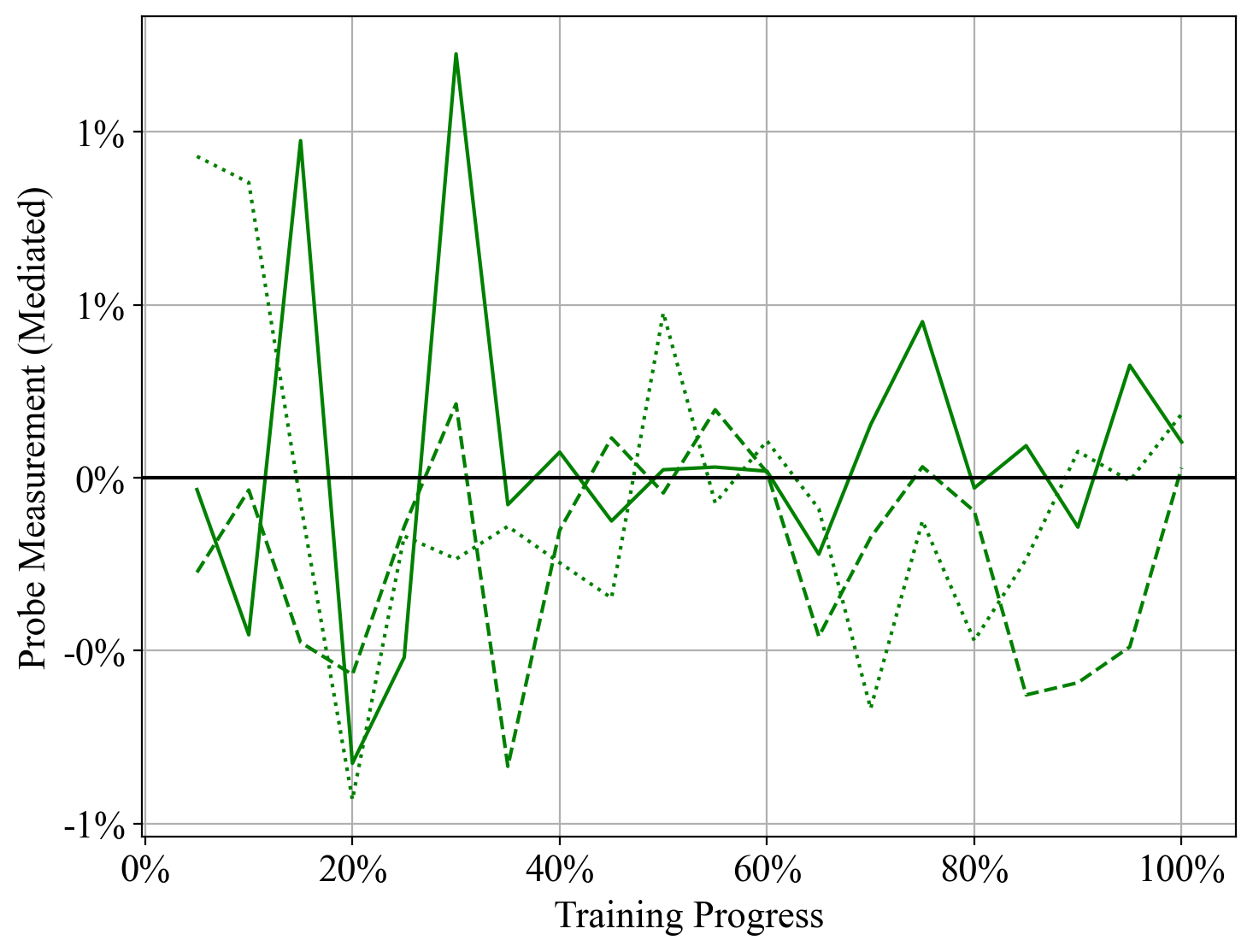}
     \caption{Deductive bias, mediated.}
     \label{fig:ia:mediated7}
 \end{subfigure}
 \hfill
 \begin{subfigure}[b]{0.48\textwidth}
     \centering
     \includegraphics[width=\textwidth]{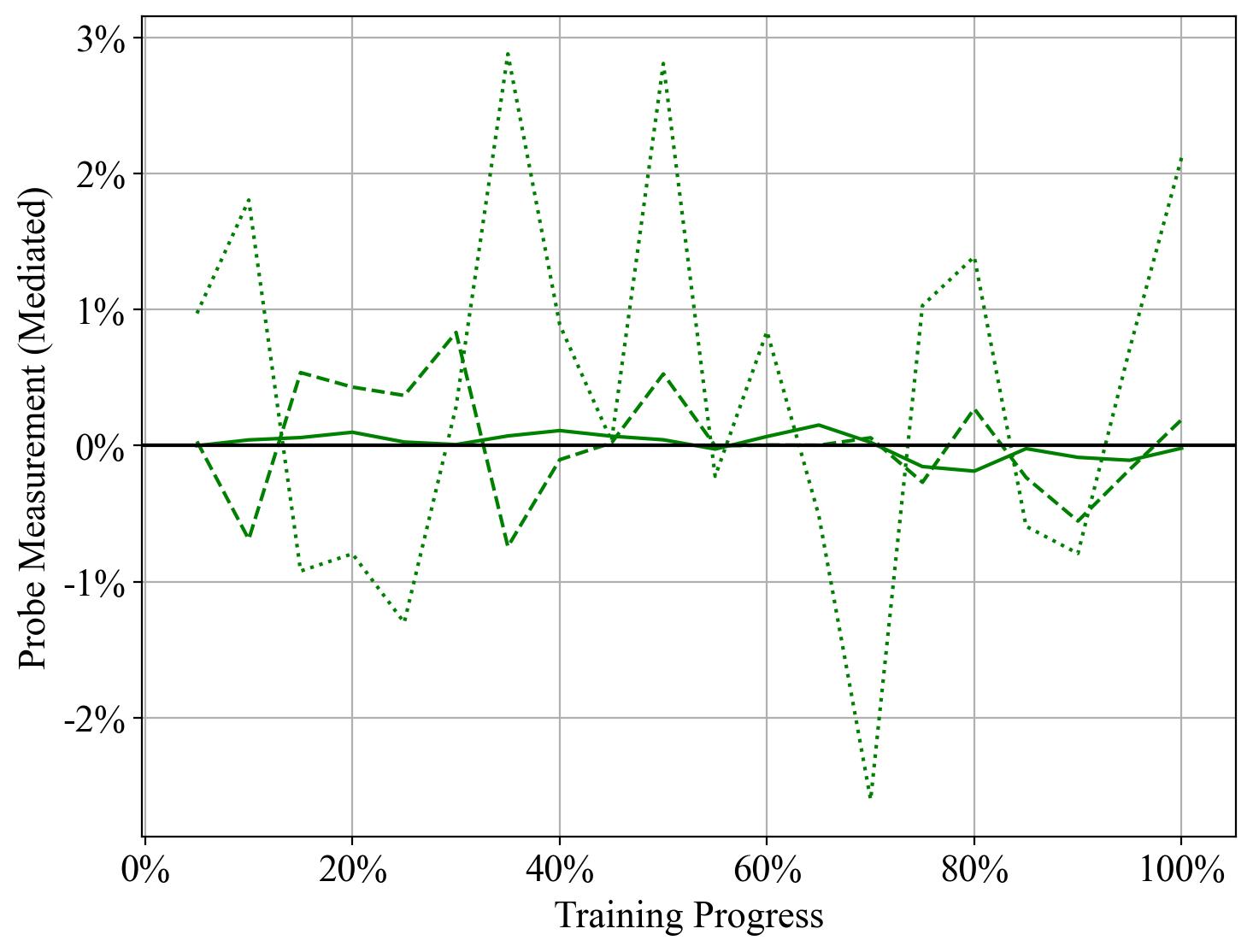}
     \caption{Inductive knowledge, mediated.}
     \label{fig:ood:mediated7}
 \end{subfigure}
\caption{Mediating with the valid baseline that swaps \texttt{turn\_right} and \texttt{turn\_left}.}
\label{fig:results_7}
\end{figure}

To test the sensitivity of the mediated results (and hence, overall conclusions) on the choice of valid baseline, we generate two additional auxiliary datasets with the following SCMs:
\begin{enumerate}
    \item swap \texttt{move} and \texttt{turn\_left}
    \item swap \texttt{turn\_right} and \texttt{turn\_left}
\end{enumerate}

The results are plotted in \Cref{fig:results_8} and \Cref{fig:results_7}, respectively. We find that the mediated measurements in the first case are nearly identical to those in \Cref{fig:results}, despite only swapping two actions (instead of permuting 3). However, in the second case, the mediated measurements are essentially noise, centered around 0. We attribute this to the fact that the resulting labels are extremely similar, as, in most cases, the robot is simply reflected along the starting axis. We emphasize that a negative result for one valid baseline does not constitute evidence to reject the hypothesis, and that a single positive result from a valid baseline is sufficient to accept the hypothesis.

\clearpage 

\subsubsection{Probe architecture and hyperparameters}

\begin{figure}[h!]
 \centering
 \begin{subfigure}[b]{0.48\textwidth}
     \centering
     \includegraphics[width=\textwidth,]{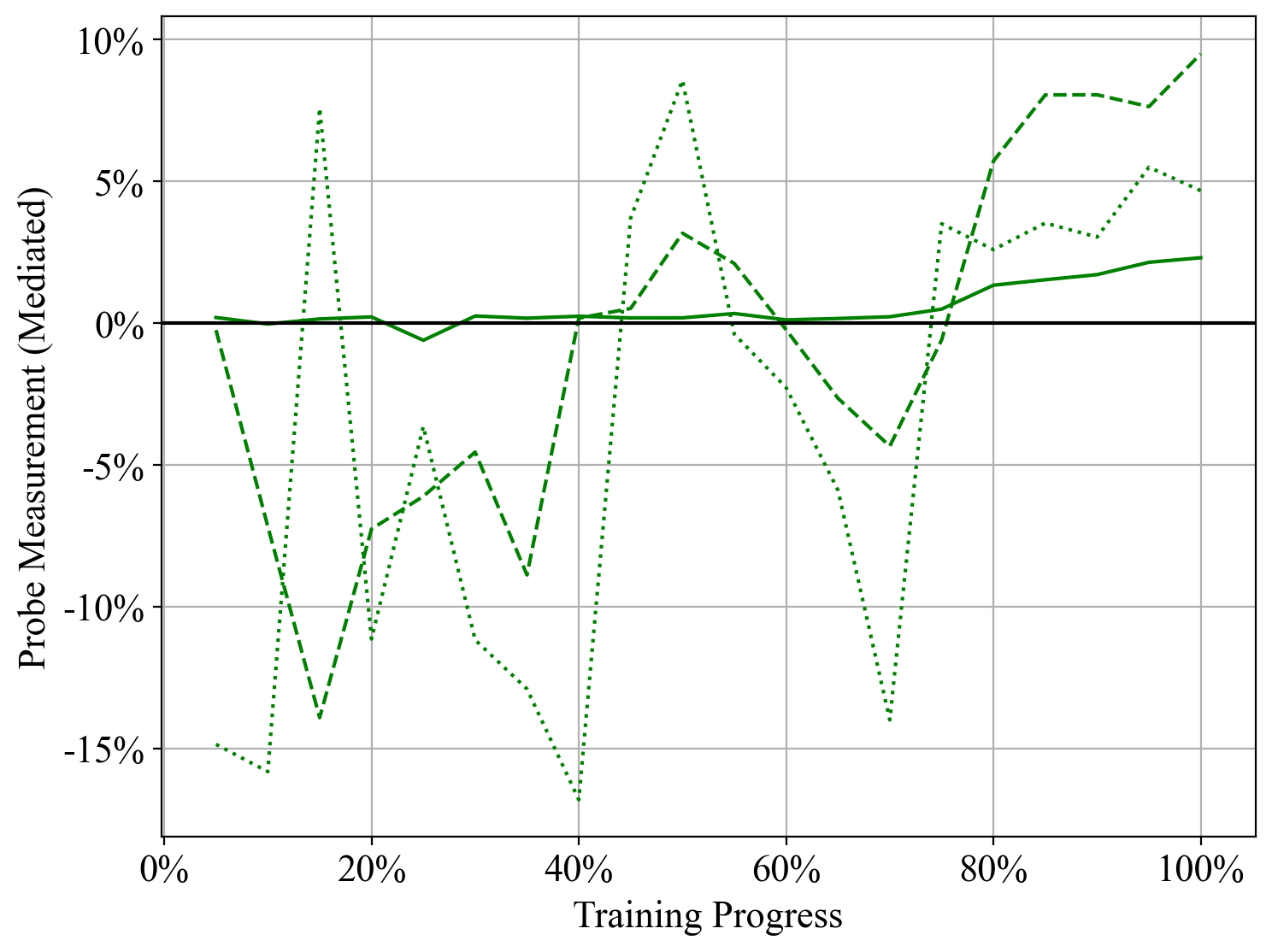}
     \caption{Deductive knowledge, mediated.}
     \label{fig:id:mediated_old}
 \end{subfigure}
 \hfill
 \begin{subfigure}[b]{0.48\textwidth}
     \centering
     \includegraphics[width=\textwidth]{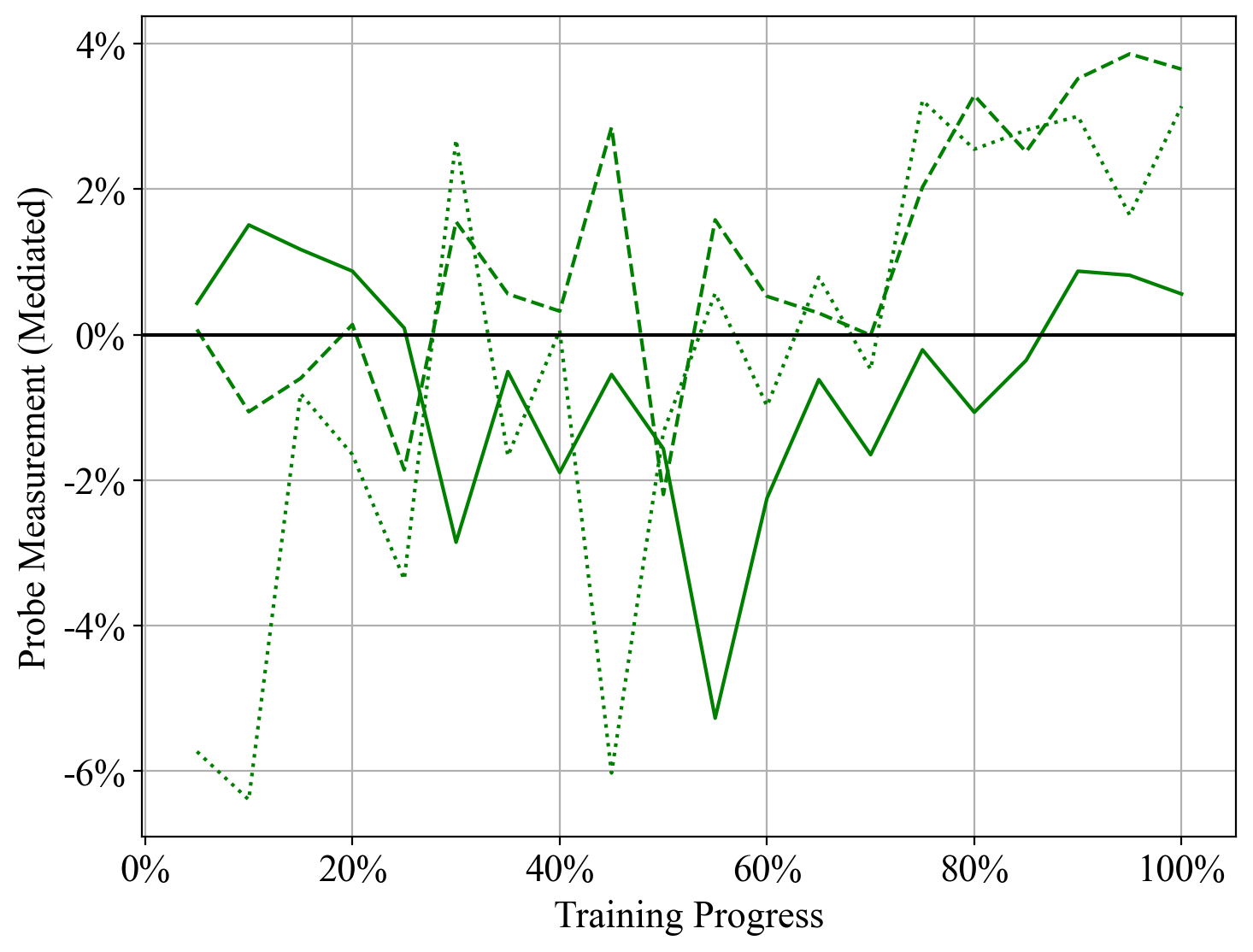}
     \caption{Inductive bias, mediated.}
     \label{fig:ig:mediated_old}
 \end{subfigure}
 \\
 \begin{subfigure}[b]{0.48\textwidth}
     \centering
     \includegraphics[width=\textwidth]{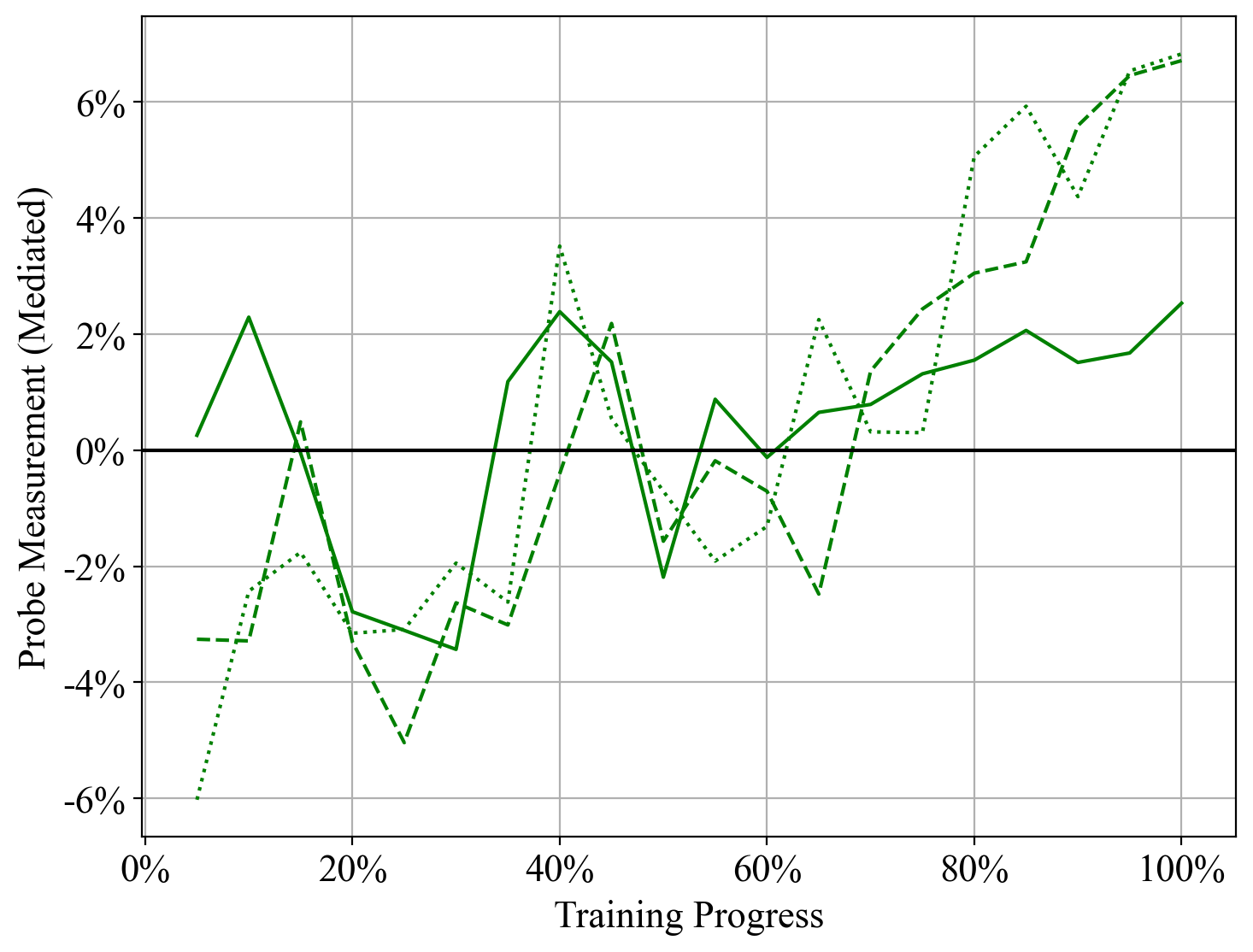}
     \caption{Deductive bias, mediated.}
     \label{fig:ia:mediated_old}
 \end{subfigure}
 \hfill
 \begin{subfigure}[b]{0.48\textwidth}
     \centering
     \includegraphics[width=\textwidth]{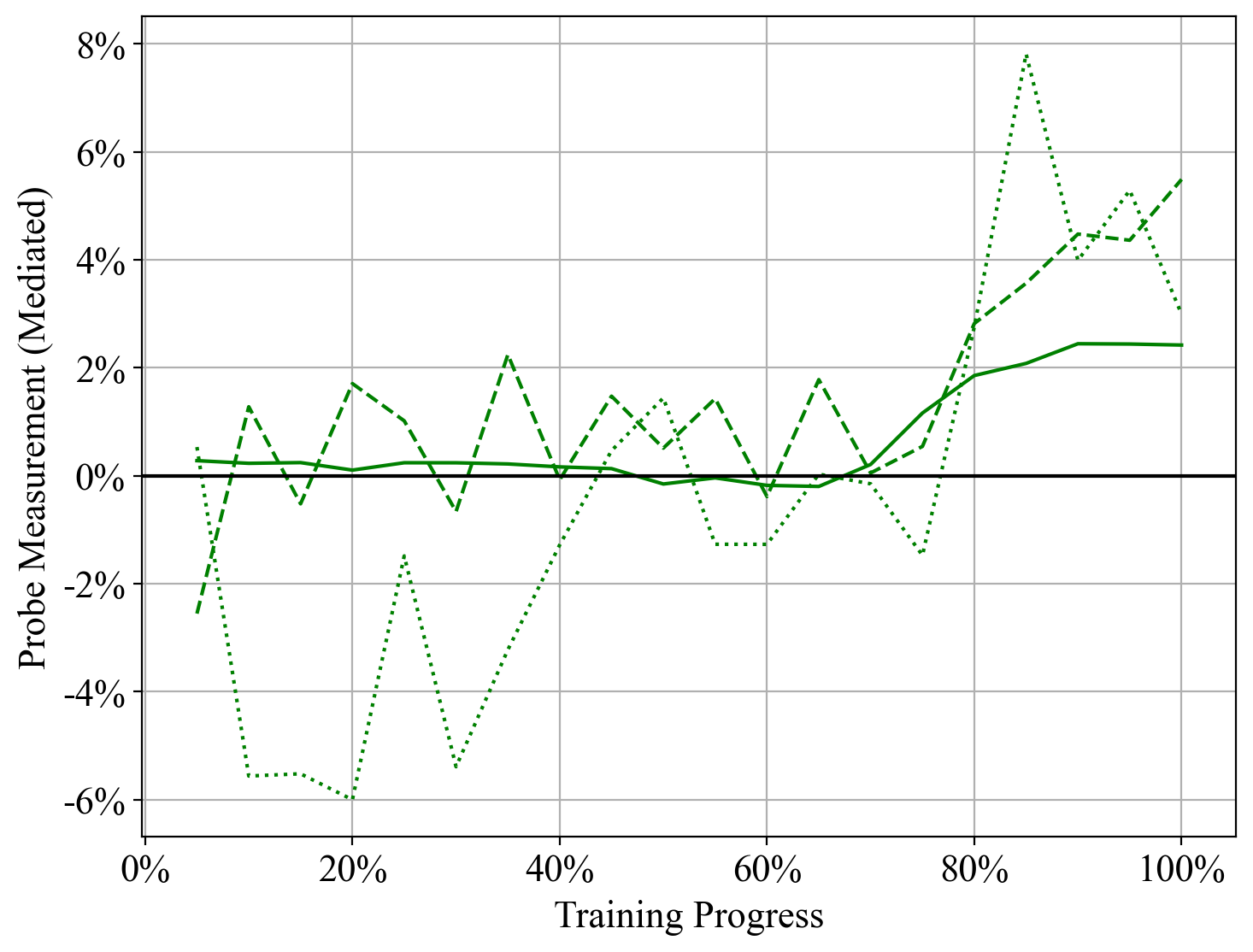}
     \caption{Inductive knowledge, mediated.}
     \label{fig:ood:mediated_old}
 \end{subfigure}
\caption{Mediating with the valid baseline from the main text. Both the original and the mediated measurements are retaken using the probe architecture and hyperparameters from \citet{jin2023evidence}.}
\label{fig:results_old}
\end{figure}

We next ablate the probe architecture and hyperparameters by adopting the settings used in \citet{jin2023evidence}. The differences are: no dropout, a batch size of 1024, training the probe for 10000000 steps, and using 100000 samples in the auxiliary dataset. We use the same valid baseline as in \Cref{fig:results} of the main text.

The results are plotted in \Cref{fig:results_old}. We observe that the general trends are preserved, and all four mediated measurements ending above 0\% by the end of training. However, we note that both deductive and inductive knowledge measure slightly lower, which is an example of the moderating effect of the probe architecture and training hyperparameters. We attribute the effect to the increased batch size and lack of dropout, which could encourage the probe to converge more quickly to a global optimum, given that the risk of overfitting is low (due to the large size and high quality of the training dataset). This is also consistent with (1) the general intuition that simpler (or less optimal) probes are a proxy for ``ease of extraction,'' which is often interpreted as evidence that the representations are ``more aligned'' with the target features \citep{hewitt2019designing}, and (2) the theoretical findings in \citet{pimentel2020information}, who conclude that probes of infinite capacity are most informative for measuring syntactic knowledge.

\clearpage

\section{Comparison with \texorpdfstring{\citet{jin2023evidence}}{Jin \& Rinard (2023)}}
\label{appendix:comp_with_icml}

In this section, we highlight several key departures from the experimental design in \citet{jin2023evidence}.

First, they do not split their auxiliary dataset into bound and free latent variable outcomes, and hence their results do not yield fine-grained interpretations about probing with different calibration and measurement datasets.

Second, our analysis reveals the presence of possible confounders in the design of their interventional baseline, leading to \emph{uncontrolled effects}. In particular, the auxiliary dataset is constructed using programs generated by the LM itself, rather than randomly sampled as we do. Intuitively, this means that the LM ``sees'' both $s_0$ and $s_n$, \emph{which reveals information about the original casual dynamics}. Formally, the representations of the LM used for probing mediates all 3 causal pathways, rather than the simple causal pathway from the LM training data (as in \Cref{fig:full_SCM}), and hence their interventional baseline is not a proper measurement of the causal effect mediated by the LM representations. Our solution is to use randomly sampled programs and replace the occurrence of $s_n$ with $s_0$ in the construction of the auxiliary dataset, which breaks this causal dependence.

Finally, \citet{jin2023evidence} do not verify that their interventional baselines satisfy the conditions in \Cref{eq:1,eq:2}. In particular, one of their baselines map the \texttt{put\_marker} and \texttt{pick\_marker} actions to \texttt{turn\_right} and \texttt{turn\_left}, respectively, in addition to permuting the \texttt{turn\_right}, \texttt{turn\_left}, and \texttt{move} actions. Because the extracted features all relate to the position and direction of the robot, the new dynamics could present a more difficult task (for both the LM and the probe) due to replacing what were effectively no-ops (\texttt{put\_marker} and \texttt{pick\_marker}) with new operations that affect the position or direction (\texttt{turn\_right}, \texttt{turn\_left}, and \texttt{move}). Hence, the observed drop in accuracy post-intervention could be attributable to increased task difficulty, rather than the learned representations of the LM.

\section{Proofs}
\label{app:proofs}

\begin{proof}[Proof of \Cref{prop:valid}]

The proof follows directly from substituting the appropriate assumptions into the definitions of NIE. Recall that
\begin{align}
\text{NIE}_{M, M'}(\theta_\text{LM}) &:= acc_{aux}(M, M) - acc_{aux}(M', M) \\
\text{NIE}_{M', M}(\theta'_\text{LM}) &:= acc_{aux}(M', M') - acc_{aux}(M, M'),
\end{align}
and, by \Cref{def:valid},
\begin{align}
\label{eq:1:app}
acc_{aux}(M', M') &\ge acc_{aux}(M, M) \\
\label{eq:2:app}
acc_{aux}(M, M') &\ge acc_{aux}(M', M).
\end{align}

Applying \Cref{eq:1:app} to the definitions of NIE,
\begin{align}
\text{NIE}_{M, M'}(\theta_\text{LM}) &\le acc_{aux}(M', M') - acc_{aux}(M', M) \\
&= \text{NIE}_{M', M}(\theta'_\text{LM}).
\end{align}

Applying \Cref{eq:2:app} to the definition of NIE,
\begin{align}
\text{NIE}_{M, M'}(\theta_\text{LM}) &= acc_{aux}(M, M) - acc_{aux}(M', M) \\
&\ge acc_{aux}(M, M) - acc_{aux}(M, M').
\end{align}

Hence,
\begin{align}
acc_{aux}(M, M) - acc_{aux}(M, M') \le \text{NIE}_{M, M'}(\theta_\text{LM}) \le \text{NIE}_{M', M}(\theta'_\text{LM})
\end{align}

\end{proof}

\end{document}